\documentclass[sigconf]{acmart}

\newif\ifarxivlayout
\arxivlayouttrue          % ← 打开以在 Overleaf 模拟 arXiv 的“更松散”排版
\ifarxivlayout
\else
\fi

\usepackage{placeins} % 可在关键处用 \FloatBarrier

\setcopyright{acmlicensed}
\copyrightyear{2025}
\acmYear{2025}
\setcopyright{cc}
\setcctype{by}
\acmConference[MM '25]{Proceedings of the 33rd ACM International Conference
on Multimedia}{October 27--31, 2025}{Dublin, Ireland}
\acmBooktitle{Proceedings of the 33rd ACM International Conference on
Multimedia (MM '25), October 27--31, 2025, Dublin,
Ireland}
\acmDOI{10.1145/3746027.3758292}
\acmISBN{979-8-4007-2035-2/2025/10}
\input{preamble}
\begin{document}

\title{MultiRef: Controllable Image Generation with Multiple Visual References}

\author{Ruoxi Chen}
\affiliation{%
  \institution{Zhejiang Wanli University}
  \city{Ningbo}
  \country{China}
}

\author{Dongping Chen\textsuperscript{\textdagger}}
\affiliation{%
  \institution{University of Washington}
  \city{Seattle}
  \country{USA}
}

\author{Siyuan Wu}
\affiliation{%
  \institution{Huazhong University of Science and Technology}
  \city{Wuhan}
  \country{China}
}

\author{Sinan Wang}
\affiliation{%
  \institution{Huazhong University of Science and Technology}
  \city{Wuhan}
  \country{China}
}

\author{Shiyun Lang}
\affiliation{%
  \institution{Huazhong
University of Science and Technology}
  \city{Wuhan}
  \country{China}
}

\author{Peter Sushko}
\affiliation{%
  \institution{Allen Institute for AI}
  \city{Seattle}
  \country{USA}
}

\author{Gaoyang Jiang}
\affiliation{%
  \institution{Huazhong University of Science and Technology}
  \city{Wuhan}
  \country{China}
}

\author{Yao Wan}
\affiliation{%
  \institution{Huazhong University of Science and Technology}
  \city{Wuhan}
  \country{China}
}

\author{Ranjay Krishna*}
\affiliation{%
  \institution{University of Washington}
  \institution{Allen Institute for AI}
  \city{Seattle}
  \country{USA}
}

\renewcommand{\shortauthors}{Chen et al.}

\begin{abstract}
Visual designers naturally draw inspiration from multiple visual references, combining diverse elements and aesthetic principles to create artwork. 
However, current image generative frameworks predominantly rely on single-source inputs - either text prompts or individual reference images.
In this paper, we focus on the task of controllable image generation using multiple visual references. We introduce \benchmark, a rigorous evaluation framework comprising 990 synthetic and 1,000 real-world samples that require incorporating visual content from multiple reference images. 
The synthetic samples are synthetically generated through our data engine \engine, with 10 reference types and 33 reference combinations. Based on \engine, we further construct a dataset \dataset containing 38k high-quality images to facilitate further research. 
% For assessment, we integrate both rule-based metrics and a fine-tuned MLLM-as-a-Judge model into \benchmark.
Our experiments across three interleaved image-text models (\emph{i.e.}, OmniGen, ACE, and Show-o) and six agentic frameworks (\emph{e.g.}, ChatDiT and LLM + SD) reveal that even state-of-the-art systems struggle with multi-reference conditioning, with the best model OmniGen achieving only 66.6\% in synthetic samples and 79.0\% in real-world cases on average compared to the golden answer. 
These findings provide valuable directions for developing more flexible and human-like creative tools that can effectively integrate multiple sources of visual inspiration. The dataset is publicly available at: \url{https://multiref.github.io/}.
\end{abstract}





\maketitle
\renewcommand{\thefootnote}{\fnsymbol{footnote}}
\footnotetext[1]{Corresponding author; \textsuperscript{\textdagger}Project leader}
\renewcommand{\thefootnote}{\arabic{footnote}}

\section{Introduction}
\label{sec:intro}
\begin{figure}
    \centering
    \includegraphics[width=0.9\linewidth]{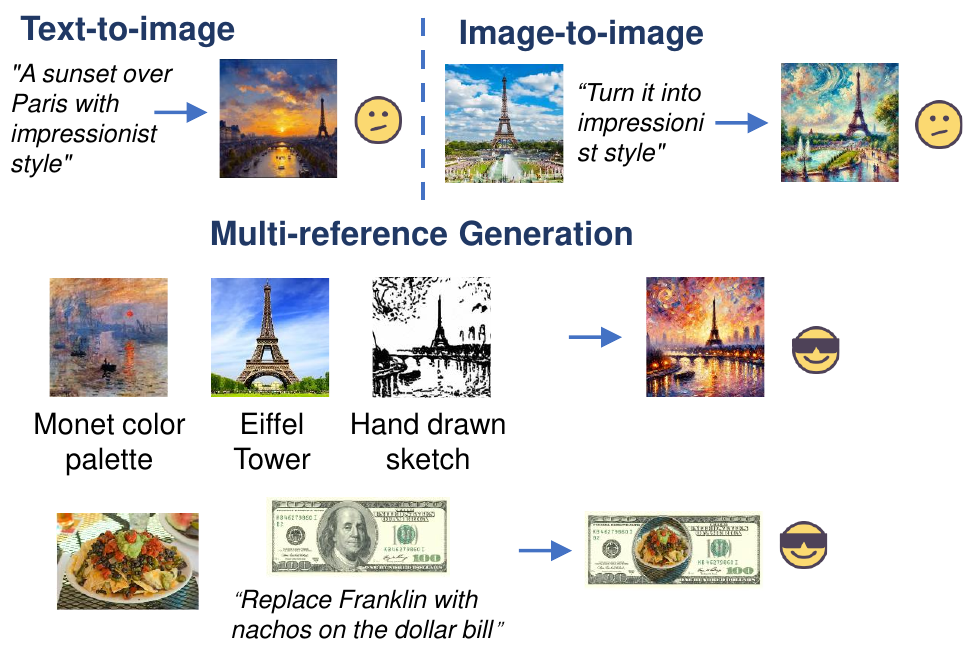}
    \vspace{-1em}
    \caption{Image generation conditioned on multiple visual references provides more controllable and creative digital art generation than single image or textual reference. 
    % \ranjay{Add more examples of multi-reference generation. Also can we call it Multi-reference Generation instead of Multi-reference-to-image?}
    }
    \vspace{-1.5em}
    \label{fig:example}
\end{figure}

Digital artists and visual designers often create a new scene by blending elements from multiple source images: a color palette from a \textit{Monet} painting, the architectural form of the \textit{Eiffel Tower} from a photograph, and the texture from a \textit{hand-drawn sketch}. Artists draw inspiration from multiple visual references, mixing diverse elements. This multi-reference creative process allows far more controllable image creation than relying on a single source of inspiration (Figure~\ref{fig:example}). However, current tools for this artistic process remain too primitive to be directly useful.
% However, tools to aid and even automate parts of this artistic process is still too primitive to be directly useful.

Despite this artistic need, today’s image generators predominantly rely on single-source conditioning—either a text prompt (\emph{i.e.}, text-to-image~\cite{rombach2022high,esser2024scaling}) or one reference image (\emph{i.e.}, image editing~\cite{ruiz2023dreambooth,li2023dreamedit}, image translation~\cite{wang2023stylediffusion,han2024stylebooth}) at a time. 
% This limitation means they struggle to emulate the flexible, multi-faceted creativity of human artists, constraining users' artistic possibilities. 
In essence, asking a modern image generative model to \textit{``paint a scene in the style of Van Gogh with the composition of a photograph''} requires specific prompt engineering~\cite{jia2024chatgen,guo2025can} or sequential editing~\cite{wang2024genartist,huang2024chatdit}.
Moreover, visual references may have inconsistent viewpoints, styles, or semantics, and merging them can produce contradictions (\emph{e.g.}, blending a daytime landscape with a night-time style reference). Existing approaches like ControlNet~\cite{zhang2023adding} excel at following one conditioning signal (\emph{i.e.}, edge map and depth), but they are not inherently designed to handle \textit{multiple} different conditions at once. 
Additionally, naively adding more control inputs usually confuses the model, leading to jumbled or degraded outputs~\cite{zhao2024uni}.

% , because the model isn’t inherently designed for image generation conditioned on multiple visual references. 

% The motivation for our work is to break this single-source barrier: to enable image generation systems that can be guided by multiple references simultaneously, offering users new levels of precision and creative control. 

% Combining several reference images (and text) into one coherent output requires precise controllability: the model must faithfully reflect each reference’s critical features. 
% , because the model isn’t inherently well trained for image generation conditioned on multiple visual references for lack of related data.

% Measuring progress towards multi-reference generation requires new benchmarks and evaluation protocols. 
Given these limitations in handling multiple references, there is a growing need to benchmark current multi-reference generation models.
From our investigation, most popular benchmarks in generative modeling focus on text-to-image alignment or single-image editing. For example, IDEA-Bench~\cite{liang2024idea} targets professional design scenarios but still typically deals with one reference at a time or sequential editing. Similarly, ACE~\cite{han2024ace} evaluates alignment with instructions but does not stress-test combining several images. No established benchmark yet examines models on truly multi-reference tasks for their integrating complexity, making it hard to quantify current research progress. 

\begin{figure*}[!t]
    \centering
    \includegraphics[width=0.8\linewidth]{ 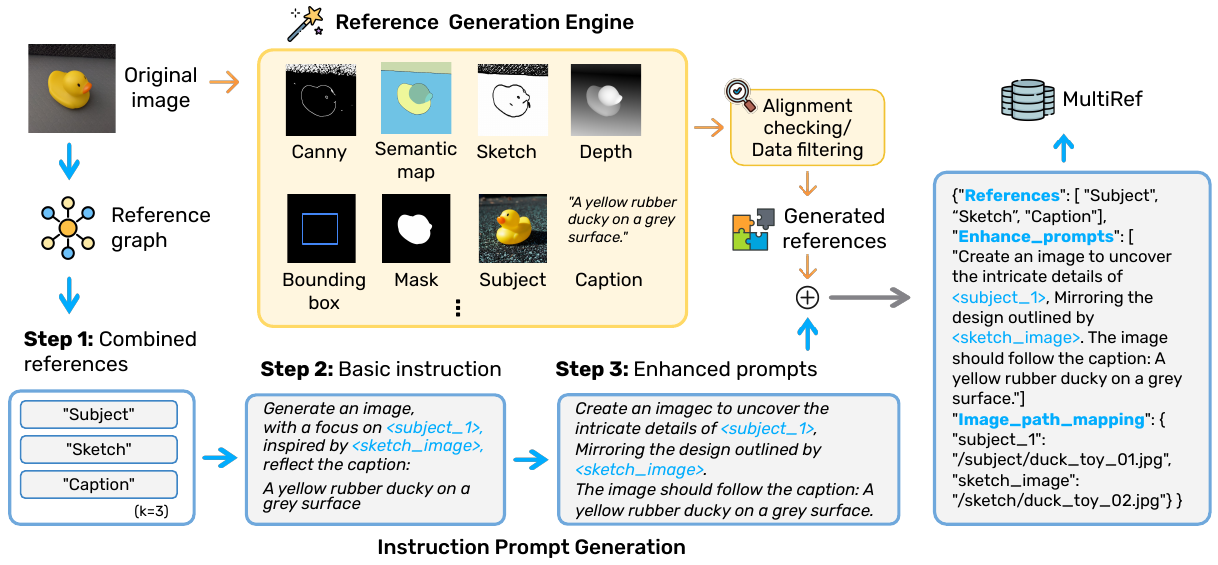}
    \vspace{-1em}
    \caption{An overview of \engine. It consists of reference generation (in yellow) and instruction prompt generation (in blue). First, various references (canny, depth, etc) are extracted from an original image. Then, a basic instruction prompt is formed from selected compatible references. Finally, the enhanced prompt is integrated with references to construct a sample.
    % \ranjay{Captions are too vague. Add more details about what the different steps are and what a reader should take away if all they did was read this caption.}
    }
    \vspace{-1em}
    \label{fig:overview}
    % \vspace{-5pt}
\end{figure*}

% To this end, this paper
In this paper, we introduce \benchmark, a benchmark that rigorously evaluates multi-reference generation models with 1,000 \emph{real-world} samples and 990 \emph{synthetic} samples that are programmatically generated. 
% For real-world samples, 
% Specifically, we assemble challenging user requests from Reddit~\cite{sushko2025realeditredditeditslargescale} where real images serve as references and golden truth to test generalization. 
Specifically, we compile challenging user requests from Reddit~\cite{sushko2025realeditredditeditslargescale}, where both references and ground truth images are real, to evaluate the image generation from multiple visual references.
% Our benchmark covers a spectrum of tasks: from relatively straightforward ones (like applying two independent style references) to complex ones requiring simultaneous spatial and semantic alignment from multiple sources. 
Our benchmark covers a spectrum of tasks, ranging from relatively straightforward scenarios---such as applying two independent references---to complex scenarios requiring simultaneous spatial and semantic alignment across multiple sources.

To address the scarcity of multi-reference image generation datasets, we develop a novel synthetic data engine, termed \engine, that efficiently creates diverse training samples. \engine first extracts various visual references (\emph{e.g.}, depth maps, Canny edge, object masks) from original images using \emph{state-of-the-art} extraction models. These references are then organized into a compatibility graph structure, where nodes represent individual references and edges indicate which references can be combined without contradictions, enabling diverse and high-quality multi-reference to image samples at scale. 
% \ranjay{This is not clear at all. What are visual reference graphs that need to be created? Why are they needed? This needs 1 or 2 sentence motivation.}
% This engine can easily curate synthetic samples by producing diverse combinations of references – \emph{e.g.}, segmentation mask + human sketch + text caption all describing aspects of the desired output – with the original image as corresponding target image. 
This engine can generate synthetic samples by flexibly combining diverse reference modalities—\emph{e.g.}, a semantic map, human pose, and caption, each describing different aspects of the intended output—while treating the original image as the corresponding target.
By controlling the data generation process, we automatically obtain rich ground-truth pairings of inputs and outputs. 
Finally, \benchmark contains 990 synthetic samples covering 10 reference types and 33 reference combinations, \emph{far surpassing} any existing collection in both scale and complexity.

We propose new protocols to evaluate the generations using our benchmark. We leverage rule-based (\emph{e.g.}, MSE for depth) and model-based (\emph{e.g.}, ClipScore~\cite{hessel2021clipscore} for aesthetic) assessments for conditions that require precise evaluation (\emph{e.g.}, depth, mask and bbox) and fine-tuned MLLM-as-a-Judge~\cite{chen2024mllm} for semantic-level assessments (\emph{e.g.}, caption, sketch and semantic map) in both reference-following and overall quality with human-annotated scores. 

% This evaluation suite not only benchmarks models’ current abilities but also provides an actionable framework for future research to track improvements in multi-reference generation. It is a first step toward standardized evaluation of controllable image generation beyond single-source inputs. 

We evaluate three interleaved image-text generation models (\emph{e.g.}, OmniGen \cite{xiao2024omnigen}, ACE \cite{han2024ace}, Show-o \cite{xie2024show}) and 6 agentic frameworks (\emph{e.g.}, ChatDiT \cite{huang2024chatdit}, LLM \cite{geminiteam2023gemini, anthropic2024claude35} + Diffusion \cite{rombach2022high,esser2024scaling}). Experimental results reveal that even the most advanced \textit{``general-purpose''} image generators today struggle with multi-reference conditioning. \emph{State-of-the-art} diffusion and autoregressive models that claim to support arbitrary conditioning (\emph{e.g.}, recent unified models) often falter when actually confronted with multiple visual inputs. 
% In our experiments, when we fed two or three reference images into , their output quality and consistency dropped noticeably – with mismatched details, missing elements, or one reference dominating the result. 
For instance, a model might capture the style of one reference image well but completely ignore the content from another subject reference.
% (\textcolor{red}{Figure~\ref{}}). 
Quantitatively, we observe substantial performance gaps: the best existing model OmniGen achieves only 0.496 of the desired alignment score on multi-reference tasks, compared to its near-perfect performance on single-reference inputs. These results expose a clear weakness in current systems – despite their advertised flexibility, they are not truly equipped for multi-reference generation. By highlighting these shortcomings, our study provides valuable insights and direction for future research. 

% It suggests that simply training on mixed modalities is not enough; new architectures or training strategies may be required to genuinely fuse information from many references. 

% By tackling the challenges of precise control and reference combination, our work lays the groundwork for next-generation controllable image generation models. The insights from our experiments – notably that current models fall short of true multi-reference agility – chart a path forward. We hope this research spurs new techniques to fuse multiple inspirations, ultimately bringing AI image generation closer to the fluid creativity of human art.

% \input{sec/2_related}
% \section{\benchmark dataset engine}
\section{\benchmark}

% \ranjay{Hey I really don't like the organization sections 3 and 4. Can you change it to the following:
% Make one big section called MultiRef-Bench. \\
% - Start out with one sentence about the overall goal of the project. \\
% - 3.1 should be a formalization of the problem. What are the inputs and outputs.\\
% - 3.2 should talk about the overall statistics of the dataset.\\
% - Use section 3.3 to go into real world examples. \\
% - 3.4 should be about data engine and synthetic examples. \\
% - 3.5 should be about the evaluation metrics}

\begin{figure*}[!t]
    \centering
    \begin{subfigure}{0.3\textwidth}
        \centering
        \includegraphics[width=\linewidth]{ 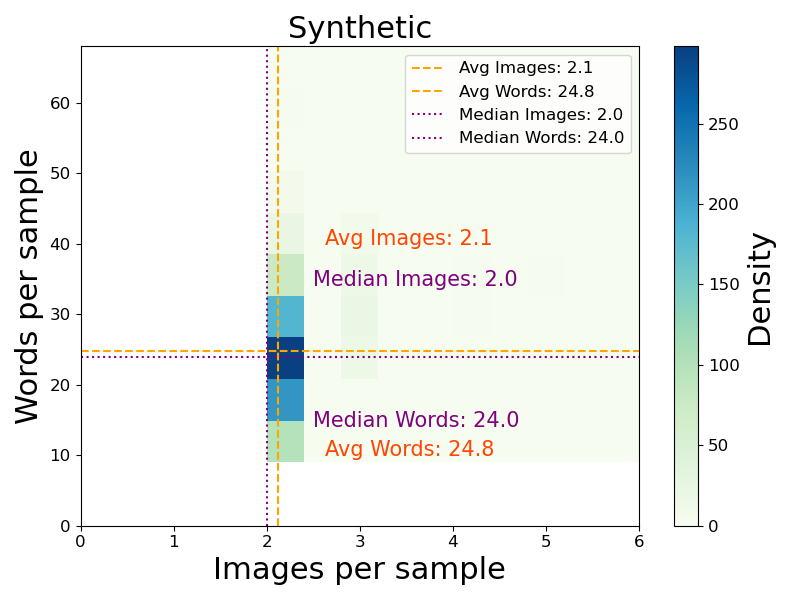}
    \end{subfigure}
    \hfill
    \begin{subfigure}{0.3\textwidth}
        \centering
        \includegraphics[width=\linewidth]{ 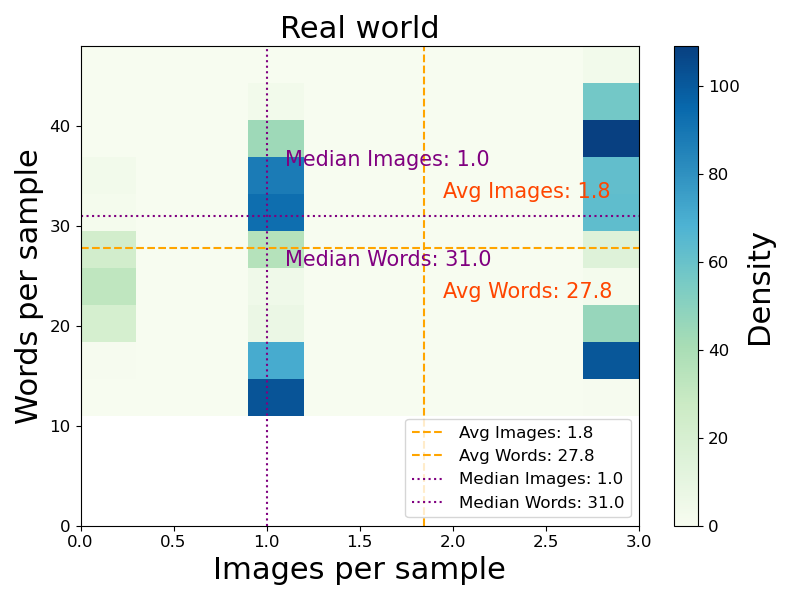}
    \end{subfigure}
    \hfill
    \begin{subfigure}{0.35\textwidth}
        \centering
        \includegraphics[width=\linewidth]{ 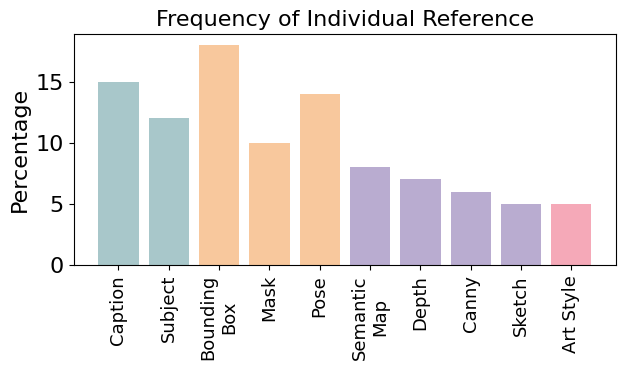}
    \end{subfigure}
    \vspace{-1em}
    \caption{\textbf{Left, Middle:} Distribution analysis of textual content length and image count for synthetic and real-world parts. \textbf{Right:} Reference frequency in synthetic data.}
    \vspace{-1em}
    \label{fig:statistic}
\end{figure*}

% \subsection{Problem Formalization}
To facilitate evaluation of multi-reference image generation models, we introduce \benchmark, the first benchmark combining real-world examples and synthetic data through a dual-pipeline methodology. The first pipeline gathers authentic tasks from publicly available sources, while the second leverages computer vision techniques to generate diverse conditional features.
\benchmark comprises 1,990 examples: 1,000 real-world tasks from Reddit's \texttt{r/PhotoshopRequest} community, selected for its diverse editing tasks and active engagement, and 990 synthetic examples from \dataset's 38,076 samples generated using \engine — our framework producing diverse guidance signals including depth maps, bounding boxes, and art styles for comprehensive conditional generation scenarios. Statistics are shown in Figure \ref{fig:statistic}.

% To facilitate the evaluation and development of image generation models with multiple reference images, we introduce \benchmark, the first benchmark of its kind. Our approach combines real-world examples and synthetic data through a dual-pipeline methodology. The first pipeline gathers real-world tasks from publicly available internet sources, capturing authentic user needs and practical challenges. The second pipeline leverages traditional computer vision techniques to generate a broad and diverse set of conditional features. By integrating these two methodologies within a single dataset, we achieve a benchmark that is not only rooted in real-world applications but also expansive, diverse, and capable of evaluating models under a wide range of possible conditions.

% \subsection{Benchmark Overview}
% \benchmark consists of 1,990 examples. The first 1,000 examples represent real-world tasks sampled from the Reddit community \texttt{r/PhotoshopRequest}. This subreddit was selected for its diverse range of editing tasks, popularity, and active user engagement.
% The remaining 990 examples are split from \dataset with 38,076 samples programmatically generated using \engine — our custom framework for generating synthetic reference images, containing a diverse set of guidance signals, including depth maps, bounding boxes, art styles, and more to produce a wide array of conditional image generation scenarios. Details regarding \benchmark statistics are shown in Figure \ref{fig:statistic}.
% \vspace{10pt}

\begin{table*}
\centering
    \caption{Reference Compatibility Matrix. Rows and columns represent reference names. \colorbox{modelcolor3}{Yellow:} Local Spatial Constraints. \colorbox{modelcolor1}{Green:} Semantic Content Specification. \colorbox{modelcolor2}{Purple:} Global Structural Guidance. \colorbox{commercialcolor}{Pink:} Semantic Content Specification. }
    \label{tab:compatibility_matrix}
    \vspace{-1em}
    \renewcommand\arraystretch{1}
     \begin{threeparttable}
    \resizebox{0.75\linewidth}{!}{
   \begin{tabular}{lcccccccccc}
   \toprule[1pt]
 & \cellcolor{modelcolor3}Bounding box & \cellcolor{modelcolor3}Mask & \cellcolor{modelcolor3}Pose & \cellcolor{modelcolor1} Caption & \cellcolor{modelcolor1} Subject & \cellcolor{modelcolor2}  Semantic map & \cellcolor{modelcolor2} Depth & \cellcolor{modelcolor2}Canny & \cellcolor{modelcolor2} Sketch & \cellcolor{commercialcolor} Art style \\
\cellcolor{modelcolor3}Bounding box & - & \xmarkcolor & \xmarkcolor & \ding{222} & \ding{222} & \xmarkcolor & \xmarkcolor & \xmarkcolor & \xmarkcolor & \checkmarkcolor \\
\cellcolor{modelcolor3}Mask & \xmarkcolor & - & \xmarkcolor & \ding{222} & \ding{222} & \xmarkcolor & \xmarkcolor & \xmarkcolor & \xmarkcolor & \checkmarkcolor \\
\cellcolor{modelcolor3}Pose & \xmarkcolor & \xmarkcolor & \checkmarkcolor & \checkmarkcolor & - & \xmarkcolor & \xmarkcolor & \xmarkcolor & \xmarkcolor & \checkmarkcolor \\
\cellcolor{modelcolor1}Caption & \checkmarkcolor & \checkmarkcolor & \checkmarkcolor & - & \checkmarkcolor & \checkmarkcolor & \checkmarkcolor & \checkmarkcolor & \checkmarkcolor & \checkmarkcolor \\
\cellcolor{modelcolor1}Subject & \checkmarkcolor & \checkmarkcolor & - & \checkmarkcolor & \checkmarkcolor & \checkmarkcolor & \checkmarkcolor & \checkmarkcolor & \checkmarkcolor & \checkmarkcolor \\
\cellcolor{modelcolor2}Semantic map & \xmarkcolor & \xmarkcolor & \xmarkcolor & \checkmarkcolor & \checkmarkcolor & - & \xmarkcolor & \xmarkcolor & \xmarkcolor & \checkmarkcolor \\
\cellcolor{modelcolor2}Depth & \xmarkcolor & \xmarkcolor & \xmarkcolor & \checkmarkcolor & \checkmarkcolor & \xmarkcolor & - & \xmarkcolor & \xmarkcolor & \checkmarkcolor \\
\cellcolor{modelcolor2}Canny & \xmarkcolor & \xmarkcolor & \xmarkcolor & \checkmarkcolor & \checkmarkcolor & \xmarkcolor & \xmarkcolor & - & \xmarkcolor & \checkmarkcolor \\
\cellcolor{modelcolor2}Sketch & \xmarkcolor & \xmarkcolor & \xmarkcolor & \checkmarkcolor & \checkmarkcolor & \xmarkcolor & \xmarkcolor & \xmarkcolor & - & \checkmarkcolor \\
\cellcolor{commercialcolor}Art style & \checkmarkcolor & \checkmarkcolor & \checkmarkcolor & \checkmarkcolor & \checkmarkcolor & \checkmarkcolor & \checkmarkcolor & \checkmarkcolor & \checkmarkcolor & - \\ \bottomrule[1pt]
\end{tabular}}
 \begin{tablenotes}
        \scriptsize
         \item[] \checkmarkcolor: possible, i.e., the combination of row and column is feasible but does not depend on each other.  \xmarkcolor: impossible, i.e., the combination of row and column is\\  invalid and cannot coexist. \ding{222}: dependency, i.e., when the row is present, the corresponding column condition must also be met.
    \end{tablenotes}
    \end{threeparttable}
\vspace{-1em}
    
\end{table*}

\subsection{Real-World Queries Collection}
To develop a robust benchmark for conditional image generation, we incorporate real-world tasks from Reddit's \texttt{r/PhotoshopRequest} community, following RealEdit~\cite{sushko2025realeditredditeditslargescale}. This platform provides authentic user-submitted editing requests requiring multiple input images.
We collected 2,300 queries, selecting tasks that combine multiple images. Each datapoint includes input images, original instructions, and output images. Manual evaluation ensures data quality by verifying image necessity, instruction coherence, and output accuracy. When multiple outputs exist, annotators select the best based on clarity and instruction fidelity.

To handle noisy instructions and specify image references, we employ GPT-4o~\cite{openai_gpt4o_2024} to generate structured prompts with image reference tokens (e.g., \texttt{<image1>}). All generated instructions undergo manual review for clarity and consistency. Missing image references are manually corrected by annotators.

\begin{table*}[t]
    \centering
      \setlength{\tabcolsep}{8pt} % Default value: 6pt

\renewcommand\arraystretch{1}
        \caption{Evaluation dimension and metrics of \benchmark for synthetic multi-ref generation. Rule: Golden standard for evaluation criteria. Model: We leverage a fine-tuned MLLM-as-a-judge for human-aligned semantic  references evaluation.}
        \label{tab:eval_metrics}
        \vspace{-1em}
 \begin{threeparttable}
    \resizebox{0.75\linewidth}{!}{
    \begin{tabular}{clllcc}
    \toprule[1.5pt]
\textbf{Evaluation Dimension} & \textbf{Evaluation Aspect} & \textbf{Evaluation Criteria} & \textbf{Quantitative Metrics} & \textbf{Rule} & \textbf{Model} \\ \hline
\multirow{2}{*}{General Quality} & Image Quality & Visual Fidelity & FID & \xmarkcolor & \ding{52} \\
 & Visual Attractiveness & Aesthetic Appeal & CLIP Aesthetic Scores & \xmarkcolor  & \ding{52} \\ \hline
 & Bounding Box & Spatial Accuracy & IoU & \ding{52} & \xmarkcolor \\
 & Semantic Map & Segmentation Accuracy & IoU & \ding{52} & \xmarkcolor \\
 & Mask & Mask Alignmen t & IoU & \ding{52}  & \xmarkcolor \\
 & Depth Map & Depth Accuracy & MSE & \ding{52} & \xmarkcolor \\
Reference Fidelity & Canny Edge & Edge Preservation & MSE & \ding{52} & \xmarkcolor \\
 & Sketch & Structural Fidelity & MSE & \ding{52} & \xmarkcolor \\
 & Caption & Text-Image Alignment & CLIP Text-Image Score & \xmarkcolor & \ding{52} \\
 & Pose \ding{93} & Pose Accuracy & mAP & \ding{52} & \xmarkcolor \\
 & Subject & Subject Consistency & CLIP Image Score & \xmarkcolor & \ding{52} \\
 & Art Style & Style Consistency & CLIP Image Score & \xmarkcolor & \ding{52} 
 \\ \midrule
 Instruction Following & Instruction Adherence & Instruction-Output Alignment & - & - & \ding{52} \\
 \bottomrule[1.5pt]
    \end{tabular}}
     \begin{tablenotes}
        \scriptsize
         \item[] \noindent \ding{93} For pose, a single reference image may contain multiple instances (e.g., multiple poses merged in one reference image). 
         % In such cases, multiple masks and multiple bounding boxes correspond to an equal number of subject references, which are provided together as input to the generation model. 
         % The evaluation metrics account for this one-to-many relationship when calculating accuracy scores.
    \end{tablenotes}
    \end{threeparttable}
    \vspace{-7pt}
\end{table*}

% \begin{figure*}[t]
%     \centering
%     \begin{subfigure}{0.3\textwidth}
%         \centering
%         \includegraphics[width=\linewidth]{ICCV2025-Author-Kit-Feb/figure/synthetic.png}
%     \end{subfigure}
%     \hfill
%     \begin{subfigure}{0.3\textwidth}
%         \centering
%         \includegraphics[width=\linewidth]{ICCV2025-Author-Kit-Feb/figure/real_world.png}
%     \end{subfigure}
%     \hfill
%     \begin{subfigure}{0.36\textwidth}
%         \centering
%         \includegraphics[width=\linewidth]{ICCV2025-Author-Kit-Feb/figure/frequency.png}
%     \end{subfigure}
%     \vspace{-5pt}
%     \caption{Left, Middle: Distribution analysis of textual content length and image count for synthetic and real-world parts. Right: Reference frequency in synthetic data (Right).}
%     \label{fig:statistic}
%     \vspace{-1em}
% \end{figure*}

We categorized datapoints using OmniEdit's taxonomy~\cite{wei2024omnieditbuildingimageediting}: Element Replacement, Element Addition, Style and Appearance Modifications, Spatial/Environment Modifications, and Attribute Transfer. Categorization was performed using GPT-4o. We balance the benchmark following the real-world distribution in the crawled raw user requests.
After rigorous quality control, 45\% of the collected data (1,000 examples) met our criteria. Each example comprises 2-6 input images, one structured instruction, and one golden output image. See Supplementary Material for more details.

\subsection{\engine: The Synthetic Data Engine}
To construct an extensive benchmark, we develop a dataset generation engine \engine\ that employs a four-step process to produce 38,076 samples across 34 reference combinations. The process includes: (1) generating reference conditioning (bounding boxes, depth maps, \emph{etc.}), (2) programmatically producing condition combinations based on compatibility rules, (3) aligning multiple references through text prompts, and (4) deploying filtering to eliminate low-quality results. 
% This approach ensures only relevant examples are included. resulting in a benchmark covering conditional image generation scenarios. To construct an extensive benchmark, we develop a custom dataset generation engine \engine, that employs a four-step process to automatically produce 38,076 diverse samples across 34 possible reference combinations. The process includes: (1) generating a comprehensive list of all potential reference conditioning (bounding boxes, depth maps, \emph{etc.}), (2) programmatically produce a unique and exhaustive set of condition combinations based on compatible rules, (3) align multiple reference though a detailed text-based prompts, and (4) deploying a high-quality filtering pipeline to eliminate low-quality results. This structured approach ensures that only the most relevant and effective examples are included in the final dataset, resulting in a diverse and robust benchmark that covers a wide range of conditional image generation scenarios.

\noindent\textbf{Step 1: Generate Reference Conditions.}
Given an original image, \engine leverages recent advanced models (\emph{e.g.} Grounded SAM2~\citep{ren2024grounded}, SAM2~\citep{ravi2024sam2}, Depth Anything2~\citep{depth_anything_v2}), to synthesize a diverse set of conditioning inputs. 
These include canny edges, semantic maps, sketches, depth maps, bounding boxes, masks, poses, art styles and subjects, along with textual captions generated by GPT-4o-mini~\citep{openai2024gpt4omini}. These reference guidance types have proved themselves in controllable image generation in prior work~\cite{zhang2023adding,zhao2024uni,qin2024unicontrol,hu2024instruct}. 

Our original images are sampled in a wide range from DreamBooth~\cite{ruiz2023dreambooth}, CustomConcept101~\cite{kumari2023multi}, Subjects200K~\cite{tan2024ominicontrol}, WikiArt~\cite{saleh2015large}, Human-Art~\cite{ju2023human}, StyleBooth~\cite{han2024stylebooth} and VITON-HD~\cite{choi2021viton}, which attach references about pose, subject, and art style within the dataset and for the diversity of metadata.

% Many references have been adopted by previous work \cite{zhang2023adding,zhao2024uni,qin2024unicontrol,hu2024instruct} for guiding image generation. We include frequently-used ones into our Reference Generation Engine: Canny edges, semantic maps, sketches, depth maps, bounding boxes, masks, poses, art styles and subjects, along with textual captions. Leveraging recent advanced models \cite{ren2024grounded,ravi2024sam2,depth_anything_v2,openai2024gpt4omini}, given an original image, we can synthesize these diverse image references.

% , such as subject-driven generation, text-guided image generation, and style transfer. Each reference type evaluates specific model capabilities: bounding boxes test spatial reasoning, depth maps assess distance perception, and semantic maps evaluate scene understanding, etc. 

\noindent\textbf{Step 2: Combining References.} References have compatibility constraints and dependencies. Some references are mutually exclusive, while others have specific dependencies that must be considered. We establish Reference Compatibility Rules that define valid combinations among different conditions, avoiding conflicts and redundancy. 
% Not all references can be combined with each other. Some references are mutually exclusive, while others have specific dependencies that must be considered. To account for these complexities, we establish a set of visual reference compatibility rules. These rules define the valid combinations and dependencies among different image reference conditions. Following the rules ensures that only non-conflicting and meaningful reference combinations are used in dataset curation, avoiding redundancy. 
Based on it, we can obtain the reference compatibility matrix for better visualization (Table \ref{tab:compatibility_matrix}). 

To ensure diversity and complexity within the dataset, we generate possible combinations of 2, 3, or 4 references per instruction while strictly adhering to compatibility rules. 
% These combinations evaluate models' capacity to integrate diverse guidance effectively.
% These combinations are designed to challenge models by requiring effective integration of multiple types of guidance.

\noindent\textbf{Visual Reference Compatibility Rules.} These rules define valid combinations among image references (style, depth, edges, segmentation), specifying compatible pairings, mutual exclusions, and dependencies.
We establish three fundamental rules for references: 
\begin{itemize} [itemsep=0pt, leftmargin=*]
    \item \textbf{Mutual Exclusivity of Global Reference:} Reference containing global information cannot be combined with each other, as this would result in information overlap. 
    %For example, Canny edge and sketch references, both containing global structural information, are mutually exclusive. 
    \item \textbf{Global-Local Information Incompatibility:} References with local information cannot be combined with those containing global information to avoid redundancy. 
    %For instance, semantic maps (global) cannot be combined with masks (local) references.
    \item \textbf{Reference Dependencies:} \ding{192} Universal Combinability: Style and caption can be combined with any other references; \ding{193} Semantic Context Requirement: Mask and bounding box references require semantic context through either subject or caption.
    % , as spatial information alone cannot specify the desired content. 
    These spatial localization references need to be attached to specific objects or concepts.
\end{itemize}

\noindent\textbf{Step 3: Generating Instructions.} Using the valid reference combinations generated in Step 2, we create two types of prompts: structured and enhanced.
Structured prompts are generated using a template-based approach that maps each reference type to a standardized phrase. For example, a depth reference might use the placeholder ``\textit{$<$depth\_image$>$}'' with associated phrases such as ``\textit{guided by the depth of $<$depth\_image$>$}.'' Caption references are appended with simple introductory phrases like ``\textit{following the caption:}''. This method ensures that prompts are clear, consistent, and easy to parse.
%, maintaining a straightforward format that models can readily interpret.

To broaden the scope and realism of our dataset, we transform structured prompts into more diverse and natural instructions using GPT-4o \cite{openai_gpt4o_2024}. By applying different personas from Persona Hub \cite{ge2024scaling}, we vary the language, tone, and style of the prompts while maintaining the reference structure and intended content. This process not only enriches the prompts with creative and contextually relevant variations but also challenges models with a wide range of linguistic expressions and scenarios. The enhanced prompts, when combined with the generated references, result in a robust and versatile dataset suitable for comprehensive model evaluation.

\noindent\textbf{Step 4: Filtering.} 
After generating visual references, we apply a rule-based filter using metrics such as a confidence score threshold of 0.8 for the IoU (Intersection over Union) of semantic maps.

For more semantic-level visual references—such as subject, style, sketch, and canny—that do not provide confidence scores, we fine-tuned Qwen-2.5-VL-7B-Instruct as a scoring-based filter \cite{chen2024mllm}. This model evaluates both the alignment between original images and generated references and their overall quality. 
Trained on 16,590 human-annotated scoring samples, the model achieves a 0.914 MAE and 0.642 Pearson correlation on a 1,750-sample test set, demonstrating performance comparable to strong proprietary models such as GPT-4o-mini. 
See Supplementary Materials for additional details.
\begin{table}[t]
\centering
\caption{Evaluating MLLM-as-a-Judge using Pearson similarity with cross-validated human-annotated ground truth. Human-Human shows the alignment between humans.}
\label{tab:small_eval_mllm_judge}
\vspace{-1em}
\resizebox{0.8\linewidth}{!}{
\begin{tabular}{l|ccc|ccc}
\toprule[1.5pt]
\multirow{2}{*}{\textbf{Model}} & \multicolumn{3}{c|}{\textbf{Realistic}} & \multicolumn{3}{c}{\textbf{Synthetic}} \\
 & IQ & IF & SF & IQ & IF & SF \\
\midrule
Gemini-2.0-Flash & 0.385 & 0.422 & 0.354 & 0.369 & 0.627 & 0.588 \\
GPT-4o-mini & \textbf{0.466} & 0.530 & 0.514 & \textbf{0.438} & 0.632 & 0.616 \\
GPT-4o & 0.432 & \textbf{0.624} & \textbf{0.613} & 0.406 & \textbf{0.668} & \textbf{0.659} \\
\rowcolor{gray!10!white} Human-Human & 0.589 & 0.665 & 0.571 & 0.629 & 0.721 & 0.694 \\
\bottomrule[1.5pt]
\end{tabular}
}
\vspace{-1em}
\end{table}
\begin{figure}[t]
    \centering
    \includegraphics[width=1\linewidth]{ 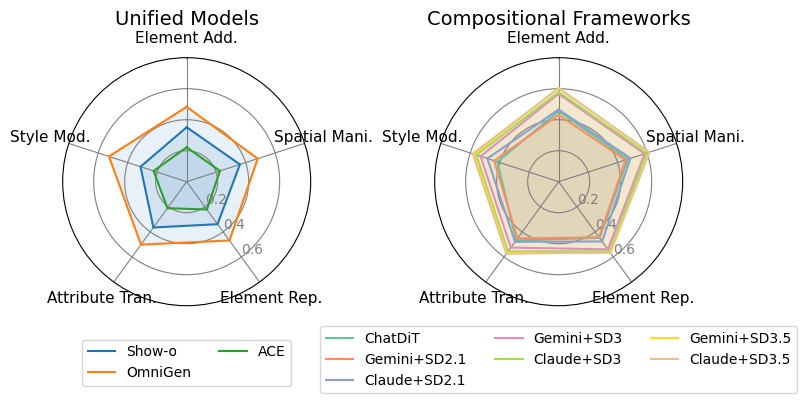}
    \vspace{-2em}
    \caption{Real-world image generation conditioned on multiple image references. Most image generative models struggle with accurately following instructions and maintaining fidelity to source images.}
    \vspace{-2em}
    \label{fig:real_radar}
\end{figure}

\begin{table*}[!t] 
\centering 
\vspace{0.5em}
\caption{Comparison of model performance on the synthetic part. Although models perform well in overall assessment, they fail at generating images with multiple references.} 
\label{tab:synth_model_comparison} 
\setlength{\tabcolsep}{3pt} % Default value: 6pt 
\vspace{-1em} 
\resizebox{0.9\textwidth}{!}{ 
\begin{tabular}{l|ccc|cc|c|cccccccccc} 
\toprule[1.5pt] 
\multirow{2}{*}{\textbf{Model}} & \multicolumn{3}{c|}{\textbf{Overall Assessment}} & \multicolumn{2}{c|}{\textbf{Image Quality}} & \multicolumn{11}{c}{\textbf{Reference Fidelity}} \\ 
% \cline{2-13} 
& IQ & IF & SF & FID$\downarrow$ & Aesthetic$\uparrow$ & AVG$\uparrow$ & BBox$\uparrow$ & Semantic Map$\uparrow$ & Mask$\uparrow$ & Depth Map$\downarrow$ & Canny Edge$\downarrow$ & Sketch$\downarrow$ & Caption$\uparrow$ & Pose*$\uparrow$ & Subject$\uparrow$ & Art Style$\uparrow$ \\ 
\midrule 
\multicolumn{17}{c}{\textcolor{violet!80!white}{\cellcolor{gray!10!white} \textbf{\textit{Unified Model}}}} \\ 
\midrule 
Show-o & \underline{0.764} & 0.616 & 0.462 & 0.110 & \underline{0.607} & 0.469 & 0.051 & 0.263 & 0.332 & 0.104 & \underline{0.061} & 0.203 & \underline{0.569} & 0.008 & 0.532 & 0.301 \\ 
OmniGen & 0.730 & 0.532 & 0.438 & \underline{0.111} & 0.593 & 0.464 & 0.179 & 0.197 & 0.320 & 0.087 & 0.092 & 0.221 & 0.382 & 0.014 & 0.623 & 0.329 \\ 
ACE & 0.740 & \underline{0.655} & \underline{0.528} & 0.108 & 0.592 & \underline{0.553} & \underline{0.219} & \underline{0.382} & \underline{0.439} & \underline{0.044} & 0.079 & \underline{0.112} & 0.521 & \underline{0.090} & \underline{0.720} & \underline{0.397} \\ 
\midrule 
\multicolumn{17}{c}{\textcolor{violet!80!white}{\cellcolor{gray!10!white} \textbf{\textit{Compositional Framework}}}} \\ 
\midrule 
ChatDiT & 0.811 & 0.713 & 0.574 & 0.100 & 0.559 & \underline{0.512} & 0.128 & \textbf{0.176} & \textbf{0.393} & 0.088 & \textbf{0.065} & \textbf{0.207} & 0.543 & \textbf{0.018} & 0.855 & 0.369 \\ 
Claude + SD 2.1 & 0.812 & 0.726 & 0.572 & \textbf{0.114} & 0.612 & 0.488 & \textbf{0.174} & 0.132 & 0.292 & 0.203 & \underline{0.080} & 0.230 & 0.547 & 0.005 & 0.817 & \underline{0.424} \\ 
Claude + SD 3 & 0.876 & 0.817 & 0.658 & 0.102 & 0.635 & 0.500 & 0.134 & 0.145 & 0.360 & 0.203 & 0.087 & 0.215 & 0.576 & \underline{0.009} & \textbf{0.859} & 0.420 \\ 
Claude + SD 3.5 & \textbf{0.913} & \textbf{0.853} & \textbf{0.691} & 0.111 & \textbf{0.647} & \textbf{0.513} & 0.124 & \underline{0.147} & 0.358 & \underline{0.082} & 0.082 & \underline{0.213} & 0.573 & \underline{0.009} & \underline{0.858} & \textbf{0.434} \\ 
Gemini + SD 2.1 & 0.791 & 0.708 & 0.547 & \underline{0.113} & 0.615 & 0.477 & \underline{0.161} & 0.133 & 0.255 & 0.202 & 0.092 & 0.239 & 0.550 & 0.003 & 0.791 & 0.406 \\ 
Gemini + SD 3 & 0.856 & 0.804 & 0.639 & 0.103 & 0.635 & 0.507 & 0.141 & 0.135 & 0.368 & 0.083 & 0.121 & 0.216 & \textbf{0.581} & 0.008 & 0.840 & 0.414 \\ 
Gemini + SD 3.5 & \underline{0.893} & \underline{0.839} & \underline{0.676} & 0.111 & \underline{0.646} & 0.510 & 0.132 & 0.130 & \underline{0.371} & \textbf{0.077} & 0.096 & 0.216 & \underline{0.579} & 0.008 & 0.845 & 0.422 \\ 
\midrule 
\rowcolor{gray!10!white} Ground Truth & 0.842 & 0.803 & 0.668 & 0.108 & 0.617 & \textbf{0.709} & \textbf{0.410} & \textbf{0.772} & \textbf{0.893} & \textbf{0.000} & \textbf{0.000} & \textbf{0.000} & \textbf{0.584} & \textbf{0.149} & \textbf{0.869} & 0.417 \\ 
\bottomrule[1.5pt] 
\end{tabular} 
} 
\vspace{-1em}
\end{table*}

% \vspace{-0.5em}
\subsection{Evaluation}

Our approach combines rule-based and model-based metrics to provide a comprehensive assessment of reference following capabilities across diverse conditions. The evaluation dimension and metrics of \benchmark are shown in Table \ref{tab:eval_metrics}. All evaluation metrics are finally normalized to a $[0,1]$ range for consistency. 
%For Reference Fidelity assessment, we calculate individual scores for each reference type, then derive the overall fidelity score by averaging across all references involved in a generation task.
% —summing the fidelity scores and dividing by the total number of references used (\textit{e.g.}, three references would yield a denominator of 3).

\noindent\textbf{Reference Fidelity.} It measures how accurately generated images preserve and incorporate the specific attributes, features, and characteristics from  reference inputs. For the 10 reference types included in our benchmark, we employ specialized evaluation criteria and metrics tailored to each reference, then derive the overall fidelity score by averaging across all references involved.
% Spatial references (Bounding Box, Semantic Map, and Mask) are evaluated using IoU to quantify alignment accuracy. For structural references (Depth map, Canny edge, and Sketch), we calculate MSE to measure preservation fidelity. Pose accuracy is quantified with mAP. Semantic references receive specialized treatment: caption alignment is assessed using CLIP text-image scores \cite{hessel2021clipscore}, while subject consistency and style fidelity are evaluated using CLIP image-image scores.
Notably, for aspects where rule-based metrics may not fully capture nuanced performance---particularly style consistency and subject fidelity---we supplement our evaluation with MLLM-as-a-Judge assessments by our finetuned model for complementary qualitative insights.

\noindent\textbf{Image Quality.} It assesses the visual quality and aesthetic appeal of generated images, independent of reference fidelity. To evaluate this dimension comprehensively, we employ two complementary metrics: FID \cite{heusel2017gans} and CLIP aesthetic scores \cite{schuhmann2022laion}, to evaluate the image quality and creative aspects of the generated content.

\noindent\textbf{Overall Assessment.} We follow \citet{chen2024mj} leveraging MLLM-as-a-Judge \cite{openai2024gpt4omini} to evaluate overall Image Quality (IQ), Instruction Following (IF), and Source fidelity (SF) in a holistic manner. We leverage GPT-4o-mini as our primary model for its superior alignment with human judgment shown in Table \ref{tab:small_eval_mllm_judge}.

\section{Experiments and Analysis}
\subsection{Experiment Setups}
% \noindent\textbf{Models.} 
We conduct evaluations on three open-source unified image generation models: OmniGen \cite{xiao2024omnigen}, ACE \cite{han2024ace}, Show-o \cite{xie2024show} \footnote{Due to computation limitation, we do not employ Emu2-Gen \cite{sheynin2024emu}.}. For ACE and Show-o, we implement multi-turn dialogues to enable image generation with multiple references, incorporating one reference image per conversational turn. 
% Additionally, we also adopt six compositional settings that leverage --- specifically Gemini-2.0-Flash \cite{geminiteam2023gemini} and Claude-3.7-Sonnet \cite{anthropic2024claude35} - as preceptors \footnote{Given that GPT-4o participated in most of our dataset filtering and benchmark construction, we do not evaluate GPT-4o \cite{openai_gpt4o_2024} here to avoid potential egocentric bias.} with dedicated image generators. SD3 \cite{esser2024scaling} as the primary generator, while using SD2.1 \cite{rombach2022high} for ablation studies. Detailed settings are in Supplementary Material.
Additionally, we evaluate six compositional settings that specifically leverage Gemini-2.0-Flash \cite{geminiteam2023gemini} and Claude-3.7-Sonnet \cite{anthropic2024claude35} as preceptors,\footnote{Given that GPT-4o participated in most of our experiments, we select alternative models for these compositional settings to avoid bias.} SD3 serves as the primary generator for dataset synthesis, with SD2.1 employed in ablation studies. See Supplementary Material for detailed configurations.

% \begin{figure}[t]
%     \centering
%     \includegraphics[width=1\linewidth]{ figure/real_radar.png}
%     \vspace{-2em}
%     \caption{Real-world image generation conditioned on multiple image references. Most image generative models struggle with accurately following instructions and maintaining fidelity to source images.}
%     \vspace{-2em}
%     \label{fig:real_radar}
% \end{figure}

\subsection{Empirical Results and Analysis}

\begin{table*}[t]
\centering
\setlength{\tabcolsep}{3pt}
% \vspace{-1em}
\caption{Ablation study on image order and caption removal using the subset of \benchmark. Switch order: switch the image input order. w/o caption: delete the caption in input instructions.}
\label{tab:ablation_order_caption}
\vspace{-1em} 
\resizebox{0.75\linewidth}{!}{
\begin{tabular}{ll|cc|cccccccccccc} % 共 15 列
\toprule[1.5pt]
\multirow{2}{*}{\textbf{Model}} & \multirow{2}{*}{\textbf{Setting}} &
\multicolumn{2}{c|}{\textbf{Image Quality}} &
\multicolumn{10}{c}{\textbf{Reference Fidelity}} \\
& & FID$\downarrow$ & Aesthetic$\uparrow$ &
BBox$\uparrow$ & Semantic Map$\uparrow$ & Mask$\uparrow$ &
Depth$\downarrow$ & Canny$\downarrow$ & Sketch$\downarrow$ &
Caption$\uparrow$ & Pose$\uparrow$ & Subject$\uparrow$ & Art Style$\uparrow$ \\
\midrule
\multirow{3}{*}{OmniGen}
& Original            & 0.114 & 0.588 & 0.267 & 0.272 & 0.273 & 0.062 & 0.098 & 0.216 & 0.014 & 0.193 & 0.735 & 0.565 \\
& Switch order   & 0.114 & 0.579 & 0.382 & 0.315 & 0.290 & 0.068 & 0.105 & 0.219 & 0.005 & 0.195 & 0.737 & 0.556 \\
& w/o caption     & 0.120 & 0.534 & 0.180 & 0.308 & 0.272 & 0.081 & 0.137 & 0.191 & -- & 0.202 & 0.669 & 0.581 \\
\midrule \midrule
\multirow{3}{*}{ACE}
& Original            & 0.114 & 0.597 & 0.326 & 0.296 & 0.311 & 0.037 & 0.089 & 0.120 & 0.089 & 0.191 & 0.715 & 0.552 \\
& Switch order   & 0.112 & 0.598 & 0.303 & 0.243 & 0.386 & 0.077 & 0.105 & 0.222 & 0.036 & 0.191 & 0.802 & 0.600 \\
& w/o caption     & 0.114 & 0.567 & 0.231 & 0.470 & 0.481 & 0.019 & 0.111 & 0.069 & -- & 0.197 & 0.657 & 0.567 \\
\midrule \midrule
\multirow{3}{*}{ChatDiT}
& Original            & 0.107 & 0.560 & 0.147 & 0.160 & 0.261 & 0.098 & 0.065 & 0.227 & 0.022 & 0.194 & 0.818 & 0.541 \\
& Switch order   & 0.105 & 0.574 & 0.125 & 0.150 & 0.284 & 0.092 & 0.063 & 0.220 & 0.022 & 0.194 & 0.830 & 0.556 \\
& w/o caption     & 0.096 & 0.550 & 0.142 & 0.132 & 0.278 & 0.113 & 0.066 & 0.196 & -- & 0.202 & 0.836 & 0.553 \\
\bottomrule[1.5pt] 
\end{tabular}
}
\end{table*}

\begin{figure*}[t]
    \centering
    \includegraphics[width=1\linewidth]{ 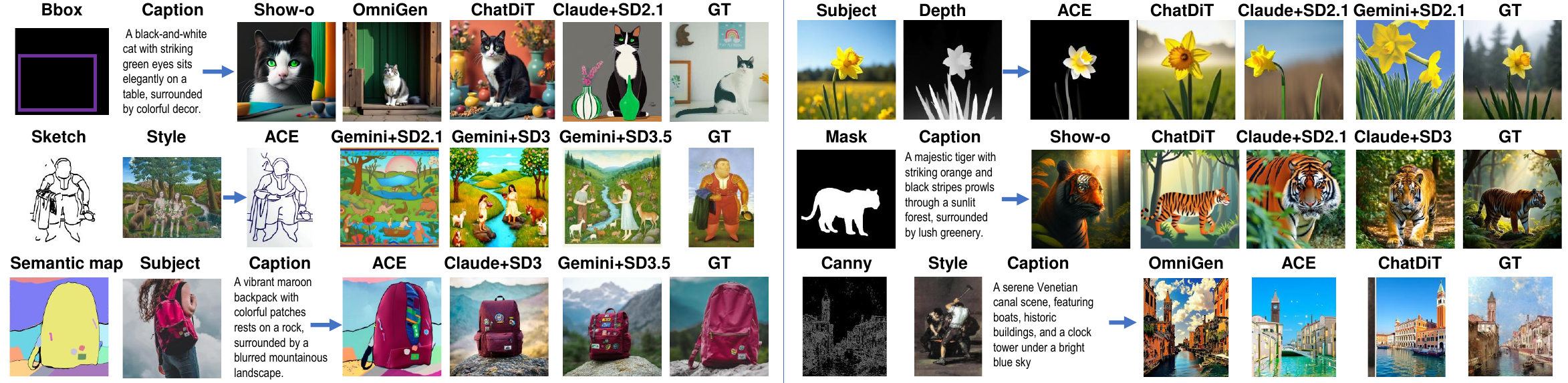}
    \vspace{-2em}
    \caption{Images generated under multiple references. More are provided in the Supplementary Materials. GT: Ground Truth.}
    \vspace{-1em}
    \label{fig:case-main}
\end{figure*}

% \begin{figure*}[t]
%     \centering
%     \includegraphics[width=0.95\linewidth]{ figure/demo-2column.pdf}
%     \vspace{-1em}
%     \caption{Case study of image generation conditioned on a combination of two and three references.
%     % \ranjay{Add more examples of multi-reference generation. Also can we call it Multi-reference Generation instead of Multi-reference-to-image?}
%     }
%     \vspace{-1em}
%     \label{fig:example1}
% \end{figure*}

\noindent\textbf{Compositional framework exceeds in image quality, while failing to maintain consistency on real-world cases.} As shown in Figure~\ref{fig:real_radar}, LLM+SD combinations achieve the highest image quality scores, with Claude + SD3.5 reaching 0.774, occasionally surpassing ground truth. However, all compositional frameworks consistently underperform in instruction following and source fidelity. While ground truth achieves 0.767 and 0.706 for IF and SF respectively, Claude + SD3.5 only reaches 0.589 and 0.462, indicating that a separated perceptor-generator architecture fundamentally compromises complex visual instruction execution.

% \noindent\textbf{Compositional framework exceeds in image quality, while failing to maintain consistency on real-world cases.} As shown in Table~\ref{fig:real_radar}, SD3.5 combined with LLMs like Gemini and Claude achieves the highest scores among all tested approaches. Claude + SD3.5 attains exceptional image quality scores of 0.774 on average, occasionally surpassing ground truth. The clear progression in scores among the three image generative models indicates that stronger image generative models achieve higher scores, demonstrating that image quality significantly impacts evaluation metrics. However, all compositional frameworks consistently underperform in accurately following users' instructions and staying fidelity to the source images in the user's query. For instance, while ground truth has 0.767 and 0.706 for IF and SF respectively, in the Overall category, Claude + SD3.5 only achieves 0.589 and 0.462, indicating that the separated preceptor-generator architecture fundamentally compromises the ability to faithfully interpret and execute complex visual instructions.

\noindent\textbf{Unified models struggled with generation quality and handling real-world images.} Although unified models theoretically end-to-end advantage that contributes to maintaining consistency, they underperform in fidelity preservation. OmniGen's performance in various metrics even approaches some compositional frameworks that generate images with \emph{state-of-the-art} diffusion models, demonstrating its effectiveness in balancing quality with instruction adherence. However, all models still fall short when compared with the golden answer (created with professional software), highlighting significant room for improvement in real-world image generation scenarios.
% while current image generative models, especially compositional frameworks, demonstrate significant capabilities in generating high-quality images, there is still a gap compared to the Ground Truth, particularly in specific tasks like Bounding Box and Semantic Map. For unified models, ACE also shows better performance in tasks like Bounding Box and Semantic Map, indicating its superior ability in these areas. The Ground Truth serves as a benchmark, with consistently high scores across all metrics. For example, FID's Ground Truth scores 0.893, significantly higher than any model. While the compositional frameworks approach these benchmarks, there is still a noticeable gap, particularly in tasks like Bounding Box and Semantic Map, indicating room for improvement in accurately following instructions and maintaining fidelity to source images.

\begin{figure}[ht]
    \centering
    \includegraphics[width=1\linewidth]{ 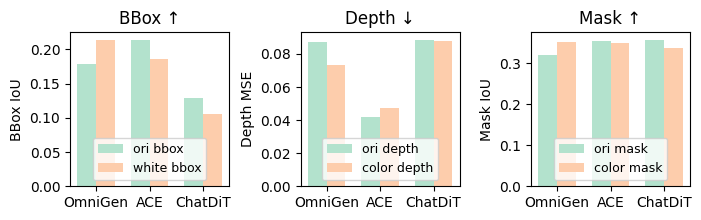}
    \vspace{-2em}
    \caption{Ablation study on the impact of different reference formats. Ori/white bbox: bounding boxes drawn in black/white backgrounds. Ori/color depth: depth maps in grey/turbo style. Ori/color mask: masks in white/light colors. }
        \vspace{-1em}
    \label{fig:condition_format_ablation}
\vspace{-1.5em}
\end{figure}

\noindent\textbf{Controllable image generation from multiple references is challenging.} Even advanced models like ACE, despite strong performance in specific areas (Bbox: 0.219, Pose: 0.090), show substantial gaps in reference fidelity compared to Ground Truth. While unified end-to-end architectures offer greater potential than compositional frameworks, both struggle with complex reference combinations or image generation without captions, highlighting the need for improved generalization in multi-image generation.

\noindent \textbf{Models show strong and varied preferences for reference formats.} Figure~\ref{fig:condition_format_ablation} presents results from an ablation study investigating how different input formats for Bounding Box (BBox), Depth, and Mask conditions affect the generation performance of three models. There is no universally superior format that works best across all tested models. Instead, each model often exhibits a distinct preference. ACE and ChatDiT show more robust performance on the depth and mask format. For Depth MSE, ACE performs significantly better with ``ori depth," whereas OmniGen and ChatDiT show slightly better or comparable performance with ``color depth".

\noindent \textbf{Input order primarily influences specific conditional fidelities rather than global image quality.} As shown in Table \ref{tab:ablation_order_caption}, switching the input order resulted in only minor FID improvements or no change for the models. However, this operation had more substantial and often model-specific impacts on adherence to particular conditions. For example, ACE's depth error and sketch error both increased dramatically when the order was switched. These observations suggest that the sequence of processing conditions is more critical for controlling specific visual attributes than for overall image realism as measured by FID. The presence of captions also improves depth fidelity and aesthetic quality across all models. For instance, ACE's depth error significantly increases when captions are removed. However, for semantic map fidelity, OmniGen and ACE perform better without captions. Similarly, sketch fidelity improves for all three models when captions are absent, with ACE showing a notable reduction in sketch error. 
% % \input{sec/5_discussion}
\section{Conclusion}
Our work presents the first investigation of image generation conditioned on multiple visual references. Through developing a sophisticated synthetic data engine, we have constructed \dataset, a large-scale dataset for multi-reference image generation, from which we carefully curated a high-quality benchmark suite alongside a real-world application to \benchmark. Our evaluation reveals that existing models face challenges when handling multi-reference generation tasks. These findings provide insights for developing models that can better handle multi-reference creative processes.
% Our work presents the first comprehensive investigation of image generation conditioned on multiple visual references, significantly expanding the boundaries of controllable image generation. Through developing a sophisticated synthetic data engine, we have constructed \dataset, a large-scale dataset for multi-reference image generation, from which we carefully curated a high-quality benchmark suite alongside a real-world application to \benchmark. Our systematic evaluation reveals that existing models, despite their claims of versatility, still face significant challenges when handling our multi-reference generation tasks. These findings provide valuable insights for the development of next-generation models that can more faithfully emulate the multi-reference creative processes inherent to human artistic expression.
% By bridging the gap between human creative processes—which naturally draw inspiration from multiple sources—and computational generation capabilities, our work establishes a foundation for more intuitive and expressive image synthesis systems that better align with artistic workflows.

\clearpage

\bibliographystyle{ACM-Reference-Format}
\bibliography{main}

% \section{Authors and Affiliations}

% Each author must be defined separately for accurate metadata
% identification.  As an exception, multiple authors may share one
% affiliation. Authors' names should not be abbreviated; use full first
% names wherever possible. Include authors' e-mail addresses whenever
% possible.

% Grouping authors' names or e-mail addresses, or providing an ``e-mail
% alias,'' as shown below, is not acceptable:
% \begin{verbatim}
%   \author{Brooke Aster, David Mehldau}
%   \email{dave,judy,steve@university.edu}
%   \email{firstname.lastname@phillips.org}
% \end{verbatim}

% The \verb|authornote| and \verb|authornotemark| commands allow a note
% to apply to multiple authors --- for example, if the first two authors
% of an article contributed equally to the work.

% If your author list is lengthy, you must define a shortened version of
% the list of authors to be used in the page headers, to prevent
% overlapping text. The following command should be placed just after
% the last \verb|\author{}| definition:
% \begin{verbatim}
%   \renewcommand{\shortauthors}{McCartney, et al.}
% \end{verbatim}
% Omitting this command will force the use of a concatenated list of all
% of the authors' names, which may result in overlapping text in the
% page headers.

% The article template's documentation, available at
% \url{https://www.acm.org/publications/proceedings-template}, has a
% complete explanation of these commands and tips for their effective
% use.
% Note that authors' addresses are mandatory for journal articles.

\section{Acknowledgments}
We would like to thank Wanting Liang, Jieyu Zhang, Weikai Huang, and Zixian Ma for their insightful feedback and support.

\clearpage

% \appendix
% This document supplements the main paper with detailed results. Below is the outline:
% \begin{itemize}
%     \item \textbf{Section 1} details the collection process of \benchmark.
%     \item \textbf{Section 2} reports the experimental setups of \benchmark.
% \end{itemize}
\appendix
\section{Related Work} 
% \header{Image Generation with Visual Reference.}

\header{Controllable Image Generation.}
The emergence of controllable image generation has revolutionized artificial intelligence by enabling users to create images that precisely match their specified criteria, including composition \cite{li2023gligen,zheng2023layoutdiffusion,yang2023reco}, style \cite{wang2023stylediffusion, ahn2024dreamstyler}, and content elements \cite{chen2024subject,chen2024anydoor}. ControlNet \cite{zhang2023adding} advanced this field by introducing spatially localized input conditions to pre-trained text-to-image diffusion models through efficient fine-tuning methods. Subsequent research \cite{ epstein2023diffusion, mou2024t2i,mo2024freecontrol,li2025controlnet} has further enhanced image controllability by implementing additional customization layers and adaptive mechanisms, enabling more sophisticated and precise image generation processes.

Building upon these advancements, some work has studied universal guidance for image generation with diffusion models \cite{bansal2023universal,xu2023versatile,pan2023kosmos,zhao2024uni,qin2024unicontrol,nair2024maxfusion,liu2024smartcontrol}. While early approaches often required complex, condition-specific adapters, a new generation of unified models has expanded possibilities by incorporating diverse input modalities to facilitate multi-modal controllable generation. These recent unified architectures support multiple visual features as conditions. Emu2-Gen \cite{sun2024generative} uses an autoregressive model to predict the next tokens and uses a separated diffusion model to generate images. 
Instruct-Imagen \cite{hu2024instruct} unifies image generation tasks together using multi-modal instructions. ACE \cite{han2024ace} introduces the condition unit designed specifically for multi-modal tasks. OmniGen \cite{xiao2024omnigen} uses an LLM as initialization and jointly models text and images within a single model to achieve unified representations across different modalities. UniReal \cite{chen2024unireal} treats image-level tasks as discontinuous video generation, enabling a wide range of image generation and editing capabilities. In parallel developments, ChatDit \cite{huang2024chatdit} employs a multi-agent system for general-purpose, and interactive visual generation.

\header{Dataset for Controllable Generation.}
Recent controllable image generation models have succeeded largely due to extensive training datasets like MultiGen-20M \cite{qin2024unicontrol}, which spans nine tasks across five categories with condition-specific instructions, while X2I dataset \cite{xiao2024omnigen} incorporates flexible multi-modal instructions - yet these approaches still predominantly address single or dual conditions rather than complex, multi-reference combinations.

% The remarkable success of controllable image generation stems largely from the extensive datasets used to train these models. 
% Several specialized datasets have emerged to support different aspects of image generation. CapsFusion-100M \cite{yu2024capsfusion} and GRIT \cite{peng2023kosmos} provide region information for text-to-image generation. Instructpix2pix \cite{brooks2023instructpix2pix} and MagicBrush \cite{zhang2024magicbrush} enable image editing, while WebLI \cite{chen2022pali} contains depth, mask and edge maps for guided generation. WikiArt \cite{saleh2015large} and StyleBooth \cite{han2024stylebooth} can be used for style transfer tasks. 
% MultiGen-20M \cite{qin2024unicontrol} spans nine tasks across five categories, pairing each condition image with specific instructions. Recent work \cite{bachmann20244m} uses pseudo labeling for multi-modal alignment, while X2I dataset \cite{xiao2024omnigen} incorporates flexible multi-modal instructions. 
% However, these datasets primarily focus on single or dual conditions, leaving more complex combinations of conditions and modalities unexplored.

Previous work has established benchmarks for evaluating image generation, primarily focused on text-to-image quality and alignment \cite{ghosh2023geneval,hu2023tifa,huang2023t2i,lin2024evaluating,gao2024generate} or image editing tasks \cite{zhang2024magicbrush,sheynin2024emu,li2023dreamedit}. Existing benchmarks like IDEA-Bench \cite{liang2024idea} and ACE benchmark \cite{han2024ace} are limited in scope, with the former including images-to-image tasks but focusing primarily on editing operations like font transfer, while the latter only evaluates alignment with textual instructions—both failing to address complex scenarios involving multiple image references and their combinations.

% \subsection{Benchmarking Image Generation}
% Previous methods have introduced benchmarks to evaluate image generation performance, primarily focusing on assessing image quality and alignment on text-to-image generation \cite{ghosh2023geneval,hu2023tifa,huang2023t2i,lin2024evaluating,gao2024generate} and image editing \cite{zhang2024magicbrush,sheynin2024emu,li2023dreamedit}. These approaches rely on single-source inputs—either text prompts or text combined with individual reference images—which inherently limits their evaluation dimensions. Evaluating models' capabilities for multi-modal reference following remains largely unexplored.
 % enabling the use of multi-image and complex instructions as diverse forms of guidance

% For generative model capabilities, IDEA-Bench \cite{liang2024idea} includes an images-to-image category, but primarily focuses on image editing tasks such as font transfer. Similarly, ACE benchmark \cite{han2024ace} only evaluates how well generated images align with the provided textual instructions. 
% Different from them,  we focus specifically on controllable image generation that enables multi-image references and complex instructions as diverse forms of generation guidance. Combining MLLM-as-a-Judge \cite{chen2024mllm} and traditional metrics, our benchmark enables a more precise and reference-specific assessment that better captures the nuances of multi-modal control.

\section{Details of Collecting \dataset}
\begin{figure}[t]
    \centering
    \includegraphics[width=0.85\linewidth]{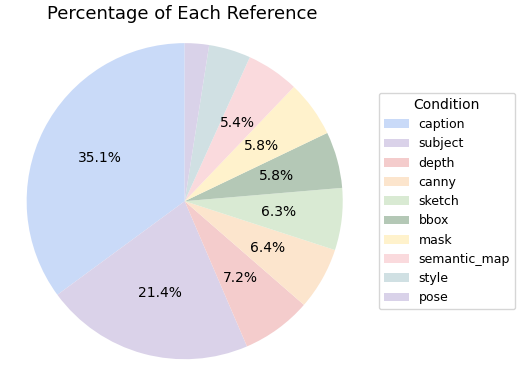}
    \caption{Reference Distribution in \dataset.}
    \label{fig:each_ref}
\end{figure}

We provide further details on the collection of \dataset. Our dataset contains 38,076 samples in total, where 990 samples are split into test set for evaluation. Using \engine, we generate more than 100k raw data, and finally gain 38k dataset of high-quality after filtering. See Table~\ref{tab:whole_data_Statistics} for detailed statistics and Figure \ref{fig:each_ref} for reference distribution in \dataset.

\begin{table}[htbp]
\centering
\caption{Distribution of Combinations by Count and Percentage in \dataset.}
\label{tab:whole_data_Statistics}
\vspace{-1em}
\resizebox{0.4\textwidth}{!}{
\begin{tabular}{lrr}
\toprule[1.5pt]
\textbf{Combination} & \textbf{Count} & \textbf{ (\%)} \\
\midrule
caption+depth+subject & 2,580 & 6.78 \\
caption+mask+subject & 2,487 & 6.53 \\
caption+depth & 2,470 & 6.49 \\
bbox+caption+subject & 2,461 & 6.46 \\
canny+caption+subject & 2,455 & 6.45 \\
caption+sketch+subject & 2,422 & 6.36 \\
canny+caption & 2,413 & 6.34 \\
caption+sketch & 2,404 & 6.31 \\
caption+semantic map+subject & 2,125 & 5.58 \\
caption+semantic map & 2,009 & 5.28 \\
bbox+caption & 1,961 & 5.15 \\
caption+mask & 1,946 & 5.11 \\
caption+pose+subject & 1,918 & 5.04 \\
depth+subject & 1,003 & 2.63 \\
caption+depth+style & 700 & 1.84 \\
canny+caption+style & 652 & 1.71 \\
caption+sketch+style & 644 & 1.69 \\
bbox+subject & 578 & 1.52 \\
bbox+caption+style & 544 & 1.43 \\
mask+subject & 542 & 1.42 \\
caption+mask+style & 535 & 1.41 \\
sketch+subject & 508 & 1.33 \\
canny+subject & 490 & 1.29 \\
caption+semantic map+style & 480 & 1.26 \\
semantic map+subject & 463 & 1.22 \\
caption+subject & 257 & 0.67 \\
pose+subject & 248 & 0.65 \\
caption+pose & 215 & 0.56 \\
canny+style & 131 & 0.34 \\
depth+style & 131 & 0.34 \\
sketch+style & 124 & 0.33 \\
semantic map+style & 93 & 0.24 \\
\midrule
\textbf{Total} & 38,076 & 100.00\% \\
\bottomrule[1.5pt]
\end{tabular}}
\vspace{-1em}
\end{table}

\subsection{Reference Generation} \label{reference_generation}
 We choose the guidance of image generation from prior work \citep{zhang2023adding,zhao2024uni,qin2024unicontrol,hu2024instruct}, combining commonly used references, standardizing their names and adapting them to work with flexible input and output formats. Our final set of references includes edge maps (Canny), semantic maps, sketches, depth maps, bounding boxes, masks, poses, art styles and subjects, along with textual captions.

\textbf{Bounding box.} A bounding box is a small possible rectangular box that can completely enclose an object in an image, typically defined by the (x,y) coordinates of its top-left and bottom-right corners. We utilize phrase grounding in Grounded SAM2 \citep{ren2024grounded} to identify and localize the main objects in a given image. The bounding box is visualized by drawing it on a black background of the same dimensions as the input image. 

\textbf{Mask.} A mask is a binary image representation where the object of interest is separated from the background. It precisely outlines the shape and contour of the target object, creating a silhouette that exactly matches the object's boundaries rather than using a rectangular bounding box. We use Grounded SAM2 to generate masks, with one object corresponding to one mask. The mask is typically visualized as a binary image, where the background is represented by black pixels (value 0), and the object mask is represented by white pixels (value 1).

\textbf{Pose.} A pose refers to the spatial arrangement of key body parts (such as head, shoulders, elbows, wrists, hips, knees, and ankles) in a human figure, typically represented as a skeleton structure with joints and connections. The pose reference is visualized on a black background, with colored joints and connections highlighting the body’s key positions and movements.

\textbf{Semantic map.} A semantic map, is a visual representation where each object class or semantic category is assigned a unique color or label, showing the location and boundaries of different semantic concepts in an image. We use AutomaticMaskGenerator in SAM2 \citep{ravi2024sam2} to generate the semantic map.

\textbf{Depth map.} A depth map is a grayscale image where each pixel's intensity represents the distance between the camera and the corresponding point in the scene. Typically, lighter/brighter pixels indicate points closer to the camera while darker pixels represent points that are farther away, creating a visual representation of the scene's 3D structure in a 2D format. We use Depth Anything V2 \citep{depth_anything_v2} to generate the depth map with default parameters.

\textbf{Canny edge.} A Canny edge map is a binary image that shows the boundaries and edges detected in an original image using the Canny edge detection algorithm. It identifies edges by looking for areas of rapid intensity change in the image, producing a clean, thin outline where white pixels represent detected edges against a black background. We use the Canny operation in OpenCV with thresholds in [100, 200].

\textbf{Sketch.} A sketch of an image is a simplified, line-based representation that captures the original image's essential contours and structural elements using only black lines on a white background. It focuses on preserving the key visual information while removing details like color, texture, and shading, similar to a hand-drawn outline. We use the line drawing method by \citet{chan2022learning} to generate the sketch reference, with contour\_style and resize\_and\_crop process. 

\textbf{Art style.} An art style of an image refers to the distinctive visual aesthetic, technique, or artistic treatment applied to transform the original image into a specific artistic rendering - such as watercolor, oil painting, cartoon and impressionist.

\textbf{Subject.} A subject reference image provides the main content or subject matter that needs to be transformed or recreated. It serves as the primary visual input that specifies what object or subject should be generated in the new image while maintaining its key characteristics and identity.

\textbf{Caption.} A caption of an image is a concise textual description that explains what is shown in the image, often describing the key subjects, actions, or notable elements present in the visual content. We use GPT-4o-mini \citep{openai2024gpt4omini} to describe the input image with prompts as follows.

\begin{tcolorbox}[prompt, title={Generate image caption}, segmentation style={draw=gray, line width=0.5mm}]
System prompt: You are a helpful assistant that can analyze images and provide detailed descriptions.\\
Here is the image: [INSERT\_IMAGES]\\

For subject-related images:\\
Describe this image in detail using no more than 20 words. Focus on the main subject in the image. Do not include any other unrelated information.
\vspace{-7pt}
\tcblower
\vspace{-7pt}
For other images:\\
Describe this image in detail using no more than 20 words. Do not include any other unrelated information.
\end{tcolorbox}

\subsection{Details of Metadata}
Original images used for reference generation are from seven datasets, as follows.

\textbf{DreamBooth \citep{ruiz2023dreambooth}.} It is a collection of images used for fine-tuning text-to-image diffusion models for subject-driven generation. It includes 30 subjects from 15 different classes. Images of the subjects are usually captured in different conditions, environments, and under different angles. While DreamBooth offers subject references, it does not include art style or pose references.

\textbf{Subjects200K \citep{tan2024ominicontrol}.} It is a large-scale dataset containing 200,000 paired images. Each image pair maintains subject consistency while presenting variations in the scene context. The dataset does not include art style or pose references. We leverage subject references provided by the dataset itself. 

\textbf{CustomConcept101 \citep{kumari2023multi}.} It is a dataset consisting of 101 concepts with 3-15 images in each concept. The categories include toys, plushies, wearables, scenes, transport vehicles, furniture, home decor items, luggage, human faces, musical instruments, rare flowers, food items, pet animals. While it offers subject references, it does not include art style or pose references.

\textbf{Human-Art \citep{ju2023human}.} It is a versatile human-centric dataset to bridge the gap between natural and artificial scenes. It includes twenty high-quality human scenes, including natural and artificial humans in both 2D representation and 3D representations. It includes 50,000 images in 20 scenarios, with annotations of human bounding box and human keypoints. From this dataset, we utilize two subsets: 2D\_virtual\_human and real\_human, containing 22,000 and 10,000 images, respectively.
Specifically, 2D\_virtual\_human provides art style and pose references while real\_human provides pose references. Additionally, we leverage the art style and pose annotations provided within the dataset.

\textbf{WikiArt \citep{saleh2015large}.} WikiArt contains art paintings from 195 different artists. The dataset has 42,129 images for training and 10,628 images for testing. It does not include the subject reference or pose reference. We use images that share the same style as the art style references. 

\textbf{StyleBooth \cite{han2024stylebooth}.} It is a high-quality style editing dataset accepting 67 prompt formats and 217 diverse content prompts, ending up with 67 different styles and 217 images per style. We use images that share the same style as the art style references.

\textbf{VITON-HD \cite{choi2021viton}.} It is a high-resolution virtual try-on dataset consisting of 11,647 person images and 11,647 corresponding clothing images. Each sample contains a person image, a target clothing image, and a pose representation. We use images as subject and pose references.

\subsection{Prompt Template}
For each reference, we generate 10 structured basic instructions, as shown below. 
% You will be given one text. please help me rewrite and polish the given text to be more diverse and creative. Do not change items in Angle brackets '<>'. Here is the text: "following the style of <style_image>" please give me ten different sentences after polishment.
% Style Conditions

\begin{tcolorbox}
[ title=Basic instructions for Art Style, enhanced,attach boxed title to top center={yshift=-3mm,yshifttext=-1mm},boxrule=0.9pt, 
  colback=gray!00,colframe=black!50,colbacktitle=TealBlue!60,
  boxed title style={size=small,colframe=TealBlue!60} ]
\begin{itemize}[leftmargin=1em] 
    \item Inspired by the essence of $\langle$style\_image$\rangle$, this reflects its distinctive flair
    \item Crafted in the characteristic tone of $\langle$style\_image$\rangle$
    \item Modeled with the unique influence of $\langle$style\_image$\rangle$
    \item Echoing the artistic spirit of $\langle$style\_image$\rangle$
    \item Infused with the signature style of $\langle$style\_image$\rangle$
    \item Reflecting the aesthetic nuances of $\langle$style\_image$\rangle$
    \item A reinterpretation influenced by $\langle$style\_image$\rangle$
    \item Harmonizing with the thematic essence of $\langle$style\_image$\rangle$
    \item Inspired by and shaped in the vein of $\langle$style\_image$\rangle$
    \item Capturing the creative vision embodied by $\langle$style\_image$\rangle$
\end{itemize}
\end{tcolorbox}

% Sketch Conditions
\begin{tcolorbox}[title=Basic instructions for Sketch, enhanced,attach boxed title to top center={yshift=-3mm,yshifttext=-1mm},boxrule=0.9pt, 
  colback=gray!00,colframe=black!50,colbacktitle=OliveGreen!60,
  boxed title style={size=small,colframe=OliveGreen!60} ]
\begin{itemize}[leftmargin=1em] 
    \item Following the sketch of $\langle$sketch\_image$\rangle$, this mirrors its essence.
    \item Designed in alignment with the sketch of $\langle$sketch\_image$\rangle$.
    \item Echoing the framework drawn by $\langle$sketch\_image$\rangle$.
    \item Guided by the outline of $\langle$sketch\_image$\rangle$, it retains its authenticity.
    \item Reflecting the initial strokes of $\langle$sketch\_image$\rangle$.
    \item Infused with the skeletal form of $\langle$sketch\_image$\rangle$.
    \item Shaped under the influence of $\langle$sketch\_image$\rangle$'s sketch.
    \item Structured around the design of $\langle$sketch\_image$\rangle$.
    \item Capturing the structural integrity of $\langle$sketch\_image$\rangle$.
    \item Crafted to reflect the framework of $\langle$sketch\_image$\rangle$.
\end{itemize}
\end{tcolorbox}

% Depth Conditions
\begin{tcolorbox}[title=Basic instructions for Depth, enhanced,attach boxed title to top center={yshift=-3mm,yshifttext=-1mm},boxrule=0.9pt, 
  colback=gray!00,colframe=black!50,colbacktitle=SeaGreen!60,
  boxed title style={size=small,colframe=SeaGreen!60}]
\begin{itemize}[leftmargin=1em] 
    \item Following the depth of $\langle$depth\_image$\rangle$, this delves into its essence.
    \item Inspired by the dimensionality of $\langle$depth\_image$\rangle$, it captures its core.
    \item Reflecting the profound layers of $\langle$depth\_image$\rangle$.
    \item Echoing the spatial depth of $\langle$depth\_image$\rangle$, it retains its integrity.
    \item Infused with the visual perspective of $\langle$depth\_image$\rangle$.
    \item Guided by the textured depth of $\langle$depth\_image$\rangle$.
    \item Structured to align with the depths captured by $\langle$depth\_image$\rangle$.
    \item Modeled after the layered depth of $\langle$depth\_image$\rangle$.
    \item Harmonizing with the multi-dimensional feel of $\langle$depth\_image$\rangle$.
    \item Crafted to embrace the depth portrayed by $\langle$depth\_image$\rangle$.
\end{itemize}
\end{tcolorbox}

% Canny Conditions
\begin{tcolorbox}[title=Basic instructions for Canny, enhanced,attach boxed title to top center={yshift=-3mm,yshifttext=-1mm},boxrule=0.9pt, 
  colback=gray!00,colframe=black!50,colbacktitle=Cerulean!40,
  boxed title style={size=small,colframe=Cerulean!40}]
\begin{itemize} [leftmargin=1em] 
    \item Following the edge of $\langle$canny\_image$\rangle$, this captures its sharpness.
    \item Inspired by the contours of $\langle$canny\_image$\rangle$, it traces its form.
    \item Reflecting the defined edges of $\langle$canny\_image$\rangle$.
    \item Echoing the precision lines of $\langle$canny\_image$\rangle$, it retains its clarity.
    \item Infused with the sharp boundaries of $\langle$canny\_image$\rangle$.
    \item Guided by the linear features of $\langle$canny\_image$\rangle$.
    \item Structured to follow the contours highlighted by $\langle$canny\_image$\rangle$.
    \item Modeled after the crisp edges of $\langle$canny\_image$\rangle$.
    \item Harmonizing with the boundary lines of $\langle$canny\_image$\rangle$.
    \item Crafted to reflect the edge details of $\langle$canny\_image$\rangle$.
\end{itemize}
\end{tcolorbox}

% Semantic Map Conditions
\begin{tcolorbox}[title=Basic instructions for Semantic Map,enhanced,attach boxed title to top center={yshift=-3mm,yshifttext=-1mm},boxrule=0.9pt, 
  colback=gray!00,colframe=black!50,colbacktitle=MidnightBlue!80,
  boxed title style={size=small,colframe=MidnightBlue!80}]
\begin{itemize}[leftmargin=1em] 
    \item Following the semantic map in $\langle$semantic\_image$\rangle$, this aligns with its meaning.
    \item Inspired by the structure of $\langle$semantic\_image$\rangle$, it conveys its intent.
    \item Reflecting the mapped semantics of $\langle$semantic\_image$\rangle$.
    \item Echoing the visual language of $\langle$semantic\_image$\rangle$, it captures its essence.
    \item Infused with the meaningful contours of $\langle$semantic\_image$\rangle$.
    \item Guided by the symbolic layout of $\langle$semantic\_image$\rangle$.
    \item Structured around the semantics depicted in $\langle$semantic\_image$\rangle$.
    \item Modeled to align with the conceptual map of $\langle$semantic\_image$\rangle$.
    \item Harmonizing with the thematic essence of $\langle$semantic\_image$\rangle$.
    \item Crafted to reflect the semantic details of $\langle$semantic\_image$\rangle$.
\end{itemize}
\end{tcolorbox}

% Bounding Box Conditions
\begin{tcolorbox}[title=Basic instructions for Bounding Box, enhanced,attach boxed title to top center={yshift=-3mm,yshifttext=-1mm},boxrule=0.9pt, 
  colback=gray!00,colframe=black!50,colbacktitle=Periwinkle,
   boxed title style={size=small,colframe=Periwinkle}]
\begin{itemize} [leftmargin=1em] 
    \item Following the bounding box in $\langle$bbox\_image$\rangle$, this outlines its structure.
    \item Inspired by the box constraints of $\langle$bbox\_image$\rangle$, it defines its scope.
    \item Reflecting the encapsulated regions of $\langle$bbox\_image$\rangle$.
    \item Echoing the boundary lines of $\langle$bbox\_image$\rangle$, it retains its precision.
    \item Infused with the spatial framework of $\langle$bbox\_image$\rangle$.
    \item Guided by the rectangular limits of $\langle$bbox\_image$\rangle$.
    \item Structured to follow the defined areas in $\langle$bbox\_image$\rangle$.
    \item Modeled after the bounding parameters of $\langle$bbox\_image$\rangle$.
    \item Harmonizing with the enclosed regions of $\langle$bbox\_image$\rangle$.
    \item Crafted to reflect the boundary specifications of $\langle$bbox\_image$\rangle$.
\end{itemize}
\end{tcolorbox}

% Single Mask Conditions
\begin{tcolorbox}[title=Basic instructions for Single Mask, enhanced,attach boxed title to top center={yshift=-3mm,yshifttext=-1mm},boxrule=0.9pt, 
  colback=gray!00,colframe=black!50,colbacktitle=Periwinkle,
  boxed title style={size=small,colframe=Periwinkle}]
\begin{itemize}[leftmargin=1em] 
    \item Following the mask in $\langle$mask\_image$\rangle$, this captures its shape.
    \item Inspired by the masked outline of $\langle$mask\_image$\rangle$, it defines its form.
    \item Reflecting the contours covered by $\langle$mask\_image$\rangle$.
    \item Echoing the masked regions of $\langle$mask\_image$\rangle$, it retains its detail.
    \item Infused with the coverage specified by $\langle$mask\_image$\rangle$.
    \item Guided by the spatial coverage of $\langle$mask\_image$\rangle$.
    \item Structured to align with the masked features in $\langle$mask\_image$\rangle$.
    \item Modeled after the outlined mask of $\langle$mask\_image$\rangle$.
    \item Harmonizing with the masked boundaries of $\langle$mask\_image$\rangle$.
    \item Crafted to reflect the regions defined by the mask in $\langle$mask\_image$\rangle$.
\end{itemize}
\end{tcolorbox}

% Pose Conditions
\begin{tcolorbox}[title=Basic instructions for Pose, enhanced,attach boxed title to top center={yshift=-3mm,yshifttext=-1mm},boxrule=0.9pt, 
  colback=gray!00,colframe=black!50,colbacktitle=RoyalPurple!50,
  boxed title style={size=small,colframe=RoyalPurple!50} ]
\begin{itemize}[leftmargin=1em] 
    \item Following the pose in $\langle$pose\_1$\rangle$, this mirrors its stance.
    \item Inspired by the posture captured in $\langle$pose\_1$\rangle$, it reflects its form.
    \item Reflecting the alignment depicted in $\langle$pose\_1$\rangle$.
    \item Echoing the position shown in $\langle$pose\_1$\rangle$, it retains its essence.
    \item Infused with the dynamic structure of $\langle$pose\_1$\rangle$.
    \item Guided by the articulated motion of $\langle$pose\_1$\rangle$.
    \item Structured around the pose outlined in $\langle$pose\_1$\rangle$.
    \item Modeled to replicate the position in $\langle$pose\_1$\rangle$.
    \item Harmonizing with the posture embodied in $\langle$pose\_1$\rangle$.
    \item Crafted to reflect the expressive pose of $\langle$pose\_1$\rangle$.
\end{itemize}
\end{tcolorbox}

% Subject Conditions
\begin{tcolorbox}[title=Basic instructions for Subject, enhanced,attach boxed title to top center={yshift=-3mm,yshifttext=-1mm},boxrule=0.9pt, 
  colback=gray!00,colframe=black!50,colbacktitle=Fuchsia!50,
  boxed title style={size=small,colframe=Fuchsia!50}]
\begin{itemize}[leftmargin=1em] 
    \item featuring $\langle$subject\_1$\rangle$.
    \item showcasing $\langle$subject\_1$\rangle$.
    \item focusing on $\langle$subject\_1$\rangle$.
    \item while emphasizing $\langle$subject\_1$\rangle$
    \item with a focus on $\langle$subject\_1$\rangle$.
    \item centered on $\langle$subject\_1$\rangle$.
    \item highlighting $\langle$subject\_1$\rangle$.
    \item to better display $\langle$subject\_1$\rangle$.
    \item while emphasizing $\langle$subject\_1$\rangle$.
    \item to reveal finer details of $\langle$subject\_1$\rangle$.
\end{itemize}
\end{tcolorbox}

We use the prompt Diversity enhancement to write enhanced instructions, shown as below.

\begin{tcolorbox}[breakable, prompt, title=Diversity enhancement]
    You will adopt the persona of {selected\_persona}. You will be given a text and your task is to rewrite and polish it in a more diverse and creative manner that reflects the persona's style. Do not include any direct references to the persona itself.\\
    You may alter sentence structure, wording, and tone.\\
    Do not modify text enclosed in angle brackets ''.\\
    If there is a 'caption:' section in the text, do not change anything following 'caption:'\\
    Here is the text: {basic\_instruction}\\
    Please provide the revised text directly without any additional commentary.

\end{tcolorbox}

\begin{table*}[htbp]
\centering
\vspace{-1em}
\caption{Basic Instructions and their enhanced prompts.}
\vspace{-1em}
\label{tab:instruction_enhance}
\begin{tabular}{c|p{.44\textwidth}|p{.44\textwidth}}
\toprule[1.5pt]
\textbf{\#} & \textbf{Basic Instruction} & \textbf{Enhanced Prompt} \\
\midrule
1 & Generate an image featuring \textcolor{cyan!60}{\texttt{<subject\_1>}}. Reflecting the encapsulated regions of \textcolor{cyan!60}{\texttt{<bbox\_image>}}. 
& 
Create an image showcasing \textcolor{cyan!60}{\texttt{<subject\_1>}}, capturing the essence of the defined areas within \textcolor{cyan!60}{\texttt{<bbox\_image>}}. 
\\
\midrule
2 & Generate an image following the edge of \textcolor{cyan!60}{\texttt{<canny\_image>}}, capturing its sharpness. Harmonizing with the thematic essence of \textcolor{cyan!60}{\texttt{<style\_image>}} following the caption: The image features a woman with a neatly styled hair bun, dressed in a simple, elegant garment. 
& 
Create an image that traces the contours of \textcolor{cyan!60}{\texttt{<canny\_image>}}, highlighting its precision. It should resonate with the thematic core of \textcolor{cyan!60}{\texttt{<style\_image>}} while adhering to the caption: The image features a woman with a neatly styled hair bun, dressed in a simple, elegant garment. 
\\
\midrule
3 & Generate an image crafted to reflect the regions defined by the mask in \textcolor{cyan!60}{\texttt{<mask\_image>}}. Following the caption: A woman in a pink dress stands confidently at a vibrant outdoor market, surrounded by colorful produce and stalls. 
& 
Generate an image designed to capture the essence of the areas delineated by the mask in \textcolor{cyan!60}{\texttt{<mask\_image>}}, following the caption: A woman in a pink dress stands confidently at a vibrant outdoor market, surrounded by colorful produce and stalls. 
\\
\bottomrule[1.5pt]
\end{tabular}
\end{table*}

Table~\ref{tab:instruction_enhance} illustrates a side-by-side comparison of the basic instructions and their enhanced versions used in our prompt design pipeline. Each basic instruction provides a concise description of the target visual generation goal, typically referencing a specific condition input (e.g., \texttt{<bbox\_image>} or \texttt{<mask\_image>}). The enhanced prompts further enrich this guidance by incorporating more descriptive verbs, clarifying intentions, and explicitly reinforcing alignment with auxiliary conditions or captions. Overall, the enhanced prompts are not only longer but also more expressive and instructive.

\subsection{Data Filtering}
To evaluate the complex outputs of free-form image generation, we assess both image quality and reference alignment using scoring-based MLLM as Filter \cite{chen2024mllm}, which has gained widespread adoption in the field \cite{liang2024idea}. 

For each reference, the multimodal large language model examines both the original and generated images, evaluating alignment between them and assessing the quality of the generated reference (if applicable). The evaluation produces numerical scores on a 5-point scale (1-5), following specific scoring rubrics detailed below.
% For semantic maps and masks, we use the logits output from SAM2 for filtering. 
References with a score under 3 will be filtered in the checking process. Pose, subject, and art style references are manually verified, as they are provided by the dataset and contain minimal annotation errors.

\begin{figure*}[h!]
  \centering
  \includegraphics[width=\textwidth]{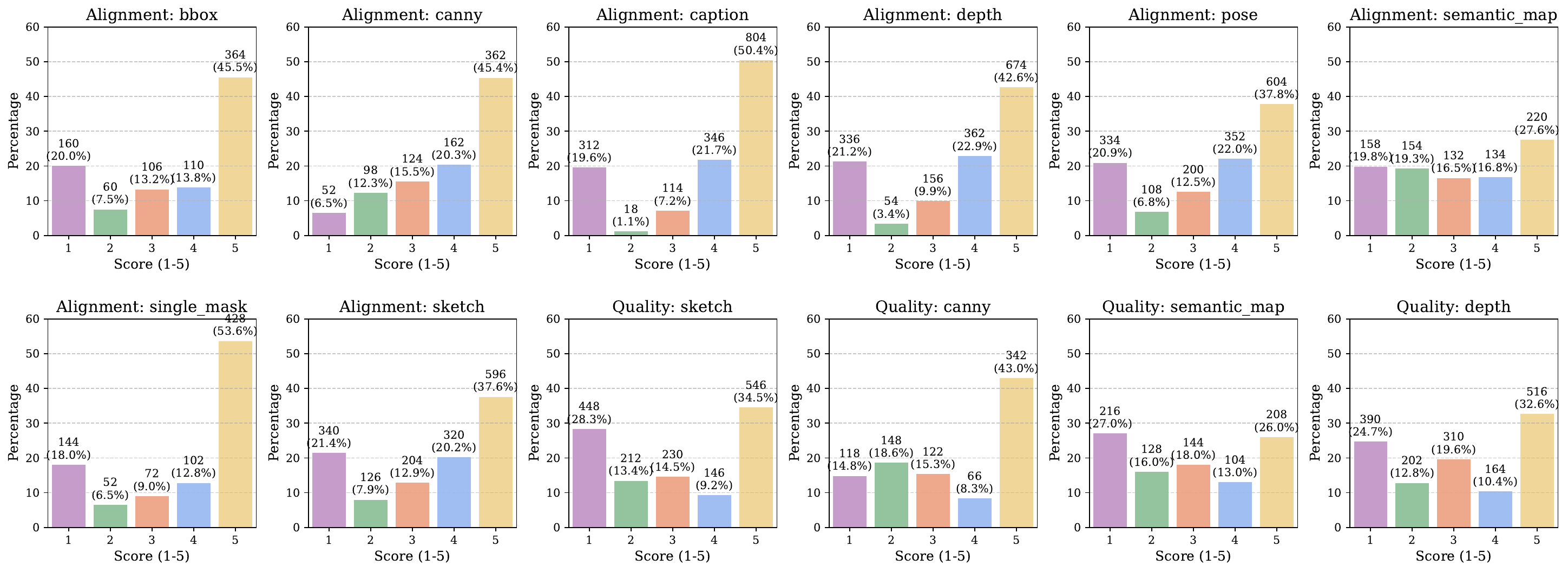} \vspace{-1.5em}
  \caption{Distribution of human preference scores (1-5) for image-text alignment and image quality across various conditioning types in the MLLM-as-a-Judge fine-tuning dataset. The top row shows alignment scores for conditions like BBox, Canny, Caption, Depth, Pose, and Semantic Map. The bottom row shows quality scores for conditions such as Single Mask, Sketch, Canny, Semantic Map, and Depth. Percentages indicate the proportion of samples receiving each score within that specific conditioning type.}
  \vspace{-1em}
  \label{fig:mllm_judge_finetune_dist}
\end{figure*}

% \subsection{Details of Fine-tuning MLLM-as-a-Judge for Filtering}
For more semantic-level visual references, such as subject, style, sketch, canny, and edge that do not provide confidence scores, we establish a fine-tuned MLLM-as-a-Judge \citep{chen2024mllm} that verifies both the alignment between original images and generated references and their quality. 
Specifically, we collect a subset with 6,400 original images and their corresponding references, constructing them into (image, reference) pairs and subsequently collect cross-validated human scoring from 1-5 for alignment and quality score individually. Finally, we split it into train/test sets, each with 16,590 and 1,750 samples, and fine-tune Qwen-2.5-7B-VL. 
Figure~\ref{fig:mllm_judge_finetune_dist} illustrates the distribution of these scores across different conditioning types and assessment criteria (alignment and quality). 

\begin{table*}[htbp]
\centering
\large

\caption{Performance comparison across different models and condition types. ZS: zero-shot, FT: finetuned.}
\vspace{-1em}
\label{tab:model_performance}
\resizebox{\linewidth}{!}{%
\begin{tabular}{l|cc|cc|cc|cc|cc|cc|cc|cc|cc|cc|cc|cc}
\toprule[1.5pt]
\multirow{2}{*}{\textbf{Model}} & \multicolumn{2}{c|}{\textbf{Overall}} & \multicolumn{2}{c|}{\textbf{Pose}} & \multicolumn{2}{c|}{\textbf{Subject}} & \multicolumn{2}{c|}{\textbf{Depth}} & \multicolumn{2}{c|}{\textbf{Caption}} & \multicolumn{2}{c|}{\textbf{Single Mask}} & \multicolumn{2}{c|}{\textbf{Style}} & \multicolumn{2}{c|}{\textbf{Sketch}} & \multicolumn{2}{c|}{\textbf{Semantic Map}} & \multicolumn{2}{c|}{\textbf{BBox}} & \multicolumn{2}{c}{\textbf{Canny}} \\
& Pearson & MAE & Pearson & MAE & Pearson & MAE & Pearson & MAE & Pearson & MAE & Pearson & MAE & Pearson & MAE & Pearson & MAE & Pearson & MAE & Pearson & MAE & Pearson & MAE \\
\midrule
Qwen2-2B-VL-ZS & 0.122 & 1.550 & 0.028 & 1.596 & 0.231 & 1.475 & 0.044 & 1.631 & 0.864 & 0.491 & 0.051 & 1.987 & 0.173 & 2.671 & 0.239 & 1.446 & 0.209 & 1.338 & -0.006 & 2.200 & 0.196 & 1.253 \\
Qwen2-2B-VL-FT & 0.602 & 0.998 & 0.102 & 1.513 & 0.787 & 0.633 & 0.695 & 0.818 & 0.874 & 0.462 & 0.308 & 1.068 & 0.473 & 0.975 & 0.617 & 1.025 & 0.497 & 1.461 & 0.093 & 1.177 & 0.270 & 1.228 \\
Qwen2-7B-VL-ZS & 0.364 & 1.397 & 0.104 & 1.603 & 0.694 & 0.892 & 0.148 & 1.757 & 0.869 & 0.515 & 0.495 & 1.757 & 0.293 & 1.171 & 0.270 & 1.515 & 0.618 & 1.117 & 0.118 & 2.588 & 0.378 & 1.309 \\
\textbf{Qwen2-7B-VL-FT (Ours)} & 0.642 & 0.914 & 0.257 & 1.397 & 0.838 & 0.557 & 0.635 & 0.894 & 0.865 & 0.485 & 0.605 & 0.797 & 0.629 & 0.728 & 0.634 & 0.946 & 0.605 & 1.162 & 0.404 & 1.129 & 0.402 & 1.124 \\
\midrule
GPT-4o-mini & 0.424 & 1.307 & 0.272 & 1.789 & 0.759 & 0.779 & 0.367 & 1.592 & 0.782 & 0.580 & 0.390 & 1.662 & 0.457 & 0.949 & 0.490 & 1.290 & 0.404 & 1.188 & 0.046 & 2.541 & 0.197 & 1.167 \\
\bottomrule[1.5pt]
\end{tabular}}
\end{table*}

Table~\ref{tab:model_performance} presents a comprehensive performance comparison of various models, evaluating their ability to align with human judgments across diverse condition types. The metrics used are Pearson correlation (higher indicates better alignment) and Mean Absolute Error (MAE, lower is better). This analysis focuses on two key aspects: the effectiveness of fine-tuning MLLM-as-a-Judge (by comparing fine-tuned "-FT" versions of Qwen models to their zero-shot base versions) and benchmarking against VQA scores \cite{lin2024evaluating} from GPT-4o-mini.

In summary, the results in Table~\ref{tab:model_performance} robustly validate the fine-tuning strategy for the MLLM-as-a-Judge, with Qwen2-7B-VL-FT achieving the best overall alignment with human preferences. While this fine-tuned judge performs well across many conditions, particularly in comparison to a general VQA model like GPT-4o-mini, challenges persist in accurately judging highly nuanced spatial conditions such as pose and bounding box accuracy. The fine-tuned model with human-annotated scores as ground truth across Pearson similarity and MAE in Table~\ref{tab:model_performance} reveals close alignment with human annotators, validating it as a good judge for filtering.

\begin{tcolorbox}[prompt, title={Eval rubrics for canny}, segmentation style={draw=gray, line width=0.5mm},breakable]

\textbf{Definitions:}\\
Canny Edge Map is a visual representation that highlights the edges and contours of objects in an image, where white lines represent detected edges and black represents non-edge regions. 
\vspace{-7pt}
\tcblower
\vspace{-7pt}
\textbf{Alignment:}
\begin{itemize}[itemsep=0pt, leftmargin=*]
    \item Not Aligned (Score 1) - Major object contours are unrecognizable or wrongly placed compared to the target image.
    \item Minimally Aligned (Score 2) - Few contours match the target image, with significant placement issues.
    \item Partially Aligned (Score 3) - Some major contours match while others are missing or misplaced.
    \item Mostly Aligned (Score 4) - Most main contours are recognizable and properly placed with minor misalignments.
    \item Well Aligned (Score 5) - Main object contours are recognizable and properly placed throughout the image.
\end{itemize}
\vspace{-7pt}
\tcbline
\vspace{-7pt}
\textbf{Quality:} 
\begin{itemize}[itemsep=0pt, leftmargin=*]
    \item Poor Quality (Score 1) - Excessive noise or breaks prevent object recognition entirely.
    \item Below Average Quality (Score 2) - Significant noise or breaks make most objects difficult to recognize.
    \item Average Quality (Score 3) - Key objects are recognizable despite moderate noise or breaks in contours.
    \item Good Quality (Score 4) - Main edges form clear object contours with minimal noise or breaks.
    \item High Quality (Score 5) - Main edges form recognizable object contours with the appropriate level of detail.
\end{itemize}
\end{tcolorbox}

% \setlength{\emergencystretch}{10pt}
% Instructions:\\
% 1. Examine the image and the generated Canny edge map.\\
% 2. Evaluate the alignment between the image and the Canny edge map.\\
% 3. Assess the quality of the Canny edge map.\\
% 4. Provide a brief explanation of your judging process.\\
% 5. Use the JSON format below to record your assessment.
% Details: \\
% \hspace*{0.4em} \text{-}\hspace*{0.2em}original image: [[original\_image\_path]] \\
% \hspace*{0.4em} \text{-}\hspace*{0.2em}Canny edge map: [[canny\_path]] \\

\begin{tcolorbox}[prompt, title={Eval rubrics for caption}, segmentation style={draw=gray, line width=0.5mm},breakable]
\textbf{Definitions:} \\
Caption is a textual description that describes the content, context, objects, actions, or scene depicted in an image. 
\vspace{-7pt}
\tcblower
\vspace{-7pt}
\textbf{Alignment:} 
\begin{itemize}[itemsep=0pt, leftmargin=*]
    \item Not Aligned (Score 1) - The caption describes elements that aren't present in the image, or fails to describe the main elements that are clearly visible.
    \item Minimally Aligned (Score 2) - The caption has minimal connection to the image content, with only one or two elements correctly identified.
    \item Partially Aligned (Score 3) -Some parts of the caption correctly describe the image while other described elements are missing or different, or the caption captures the general scene but misses key elements.
    \item Mostly Aligned (Score 4) - The caption describes most main elements and the overall scene with minor inaccuracies or omissions.
    \item Well Aligned (Score 5) - The caption accurately describes the main elements and scene in the image.
\end{itemize}

% \setlength{\emergencystretch}{10pt}
% Instructions:\\
% 1. Examine the image and the text caption. \\
% 2. Evaluate the alignment between the image and the text caption. \\
% 3. Provide a brief explanation of your judging process. \\
% 4. Use the JSON format below to record your assessment. 
% Details: \\
% \hspace*{0.4em} \text{-}\hspace*{0.2em}original image: [[original\_image\_path]] \\
% \hspace*{0.4em} \text{-}\hspace*{0.2em}caption: [[caption]] \\

%Response Format: \\
%\{
%``caption\_align\_answer'': ``\textless Align/Partially Align/Not Align\textgreater '',\\
%``thinking\_process'': ``\textless Explanation of your analysis\textgreater''\}
\end{tcolorbox}

\begin{tcolorbox}[prompt, title={Eval rubrics for sketch}, segmentation style={draw=gray, line width=0.5mm},breakable]
\textbf{Definitions:} \\
A sketch is a simplified, hand-drawn representation of an image, typically in black and white or grayscale, focusing on the main outlines and shapes of objects. 
\vspace{-7pt}
\tcblower
\vspace{-7pt}
\textbf{Alignment: }
%\hspace*{0.4em} \text{-}\hspace*{0.2em}``Align'' the basic shapes and composition match (proportions can vary, textures and details not required, stylistic interpretations acceptable).\\
%\hspace*{0.4em} \text{-}\hspace*{0.2em}``Partially Align''  if the main concept is recognizable but with significant structural deviations.\\
%\hspace*{0.4em} \text{-}\hspace*{0.2em}``Not Align'' if the basic object or scene structure is not captured.\\
\begin{itemize}[itemsep=0pt, leftmargin=*]
    \item Not Aligned (Score 1) - The basic object or scene structure is not captured at all.
    \item Minimally Aligned (Score 2) - Vague resemblance to the original image with major structural inaccuracies.
    \item Partially Aligned (Score 3) -The main concept is recognizable but with significant structural deviations.
    \item Mostly Aligned (Score 4) - Basic shapes and composition generally match with minor proportional variations.
    \item Well Aligned (Score 5) - The basic shapes and composition match accurately to the original image.
\end{itemize}
\vspace{-7pt}
\tcbline
\vspace{-7pt}
\textbf{Quality:} 
\begin{itemize}[itemsep=0pt, leftmargin=*]
    \item Poor Quality (Score 1) - Excessive noise or unclear lines make it difficult to interpret the intended subject.
    \item Below Average Quality (Score 2) - Substantial noise or rough elements that significantly detract from the subject.
    \item Average Quality (Score 3) -The sketch shows the subject but includes noticeable noise, scattered marks, or rough elements while maintaining recognizable forms.
    \item Good Quality (Score 4) - Clear lines with minimal noise that effectively represent the subject.
    \item High Quality (Score 5) - Clean, clear lines that effectively convey the subject with minimal noise or distraction.
\end{itemize}

%Sketch Quality evaluates how clean and clear the sketch is in representing the subject. Quality is considered: \\
%\hspace*{0.4em} \text{-}\hspace*{0.2em}High: if the sketch has clean, clear lines that effectively convey the subject with minimal noise or distraction \\
%\hspace*{0.4em} \text{-}\hspace*{0.2em}Average: if the sketch shows the subject but includes noticeable noise, scattered marks, or rough elements while maintaining recognizable forms \\
%\hspace*{0.4em} \text{-}\hspace*{0.2em}Poor: if excessive noise or unclear lines make it difficult to interpret the intended subject \\

% \setlength{\emergencystretch}{10pt}
% Instructions:\\
% 1. Examine the image and the generated sketch.\\
% 2. Evaluate the alignment between the image and the sketch.\\
% 3. Assess the quality of the sketch.\\
% 4. Provide a brief explanation of your judging process.\\
% 5. Use the JSON format below to record your assessment.
% Details: \\
% \hspace*{0.4em} \text{-}\hspace*{0.2em}original image: [[original\_image\_path]] \\
% \hspace*{0.4em} \text{-}\hspace*{0.2em}sketch: [[sketch\_path]] \\

%Response Format: \\
%\{
%``sketch\_align\_answer'': ``\textless Align/Partially Align/Not Align\textgreater '',\\
%``sketch\_quality\_answer'': ``\textless High/Average/Poor\textgreater '',\\
%``thinking\_process'': ``\textless Explanation of your analysis\textgreater''
%\}
\end{tcolorbox}

\begin{tcolorbox}[prompt, title={Eval rubrics for semantic map}, segmentation style={draw=gray, line width=0.5mm},breakable]

\textbf{Definitions:}\\
A semantic map is a visual representation where an image is divided into distinct regions to represent different objects, areas, or elements of the scene, using any colors or styles to distinguish between regions.
\vspace{-7pt}
\tcblower
\vspace{-7pt}
Alignment:
\begin{itemize}[itemsep=0pt, leftmargin=*]
    \item Not Aligned (Score 1) - The basic scene structure or main objects are unrecognizable.
    \item Minimally Aligned (Score 2) - Only a few elements are recognizable, with significant missing or misplaced components.
    \item Partially Aligned (Score 3) - Some key elements are recognizable but others are missing or unclear.
    \item Mostly Aligned (Score 4) - Most elements capture recognizable objects and scene layout with minor inaccuracies.
    \item Well Aligned (Score 5) -The map captures recognizable objects and scene layout appropriately (simplified shapes are acceptable, textures and fine details not required).
\end{itemize}
\vspace{-7pt}
\tcbline
\vspace{-7pt}
\textbf{Quality: }
\begin{itemize}[itemsep=0pt, leftmargin=*]
    \item Poor Quality (Score 1) - Semantic regions are too sparse or scattered to identify main objects; regions are too minimal to understand scene content.
    \item Below Average Quality (Score 2) - Main elements are barely distinguishable with significant noise, artifacts, or fragmented segments that impair understanding.
    \item Average Quality (Score 3) - Main elements are clearly visible but with noticeable noise/artifacts or scattered segments, while still maintaining recognizable object shapes.
    \item Good Quality (Score 4) - Key objects/regions are well-defined with limited noise or artifacts; segmentation is generally clean with only minor issues.
    \item High Quality (Score 5) - Main objects/regions are clearly visible and distinguishable, with clean segmentation of major elements; minimal artifacts or noise around edges.
\end{itemize}
\end{tcolorbox}

\begin{tcolorbox}[prompt, title={Eval rubrics for mask}, segmentation style={draw=gray, line width=0.5mm},breakable]

\textbf{Definitions:}\\
Mask Image is a binary image where white regions indicate areas of interest or target regions for object placement/generation, while black regions represent background or non-target areas.
\vspace{-7pt}
\tcblower
\vspace{-7pt}
\textbf{Alignment:}
\begin{itemize}[itemsep=0pt, leftmargin=*]
    \item Not Aligned (Score 1) - Main parts of the main object are not covered by the mask, or the mask position doesn't correspond to the object location.
    \item Minimally Aligned (Score 2) - The mask covers only a small portion of the main object or is significantly misplaced.
    \item Partially Aligned (Score 3) - The mask covers most but not all of the main object, or if positioning is noticeably off.
    \item Mostly Aligned (Score 4) - The mask covers the main object with minor positioning issues or slight shape inaccuracies.
    \item Well Aligned (Score 5) - The mask captures the general outline and position of the main object accurately.
\end{itemize}
\end{tcolorbox}

\subsection{Human Annotation}

The annotation process was conducted by three independent evaluators: two authors of this paper and one volunteer. Recognizing that annotator diversity is essential for minimizing bias and maximizing dataset reliability, we selected annotators with varying demographic characteristics (gender, age, and educational background) while ensuring all possessed domain expertise in image generation evaluation.

To establish annotation consistency and objectivity, all evaluators underwent comprehensive training sessions before beginning the task. These sessions included detailed tutorials on objective image assessment techniques, familiarization with reference rubrics, and instruction on the specific criteria used in our Score Evaluation framework.

\begin{figure}[h]
    \centering
    \includegraphics[width=1.0\linewidth]{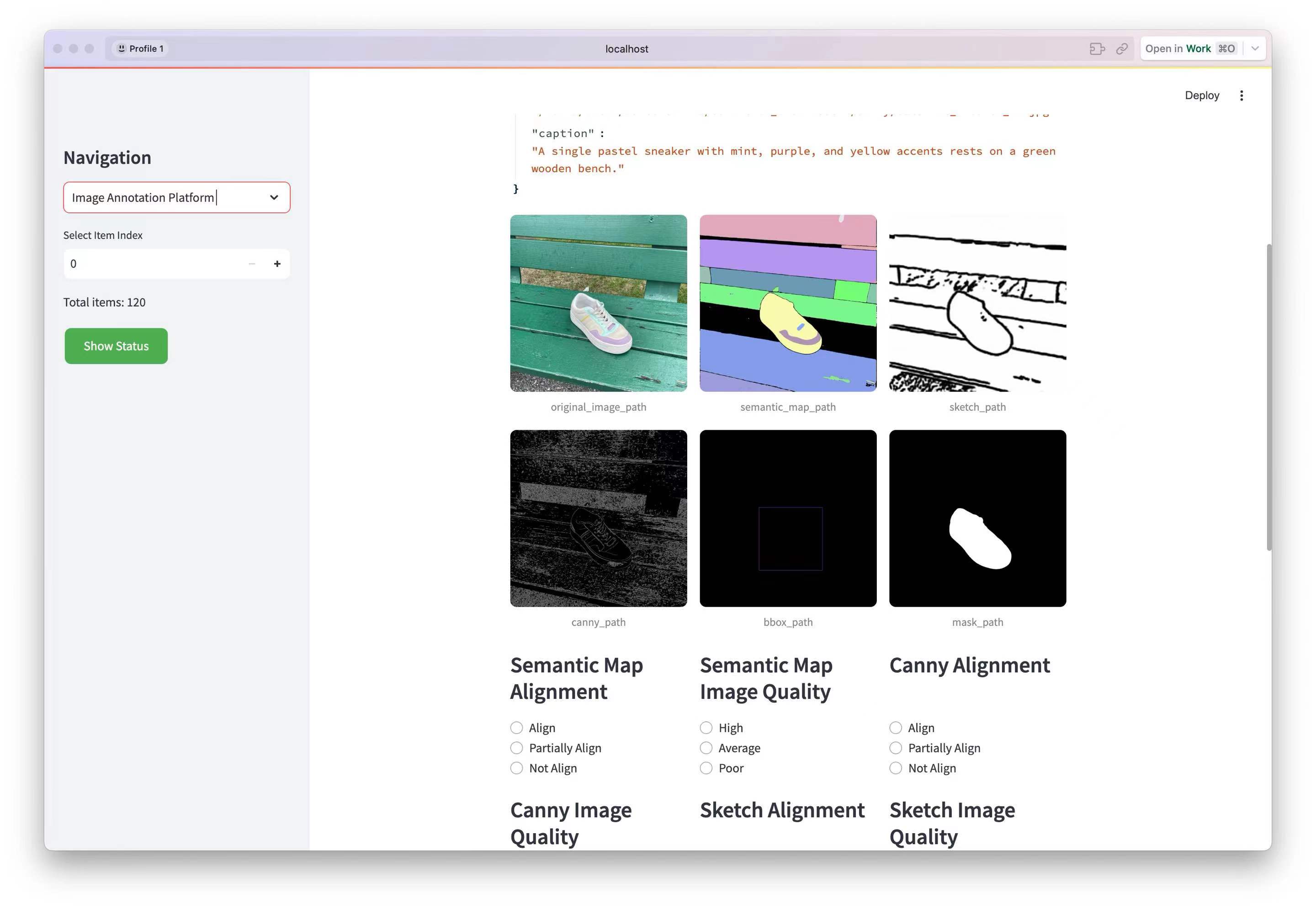}
    \vspace{-1em}
    \caption{Human Annotation Platform}
    \vspace{-1em}
    \label{fig:annotation_platform}
\end{figure}

The annotation platform is shown in Figure \ref{fig:annotation_platform}.

\section{Details of \benchmark}
\begin{figure*}
    \centering
    \includegraphics[width=0.8\linewidth]{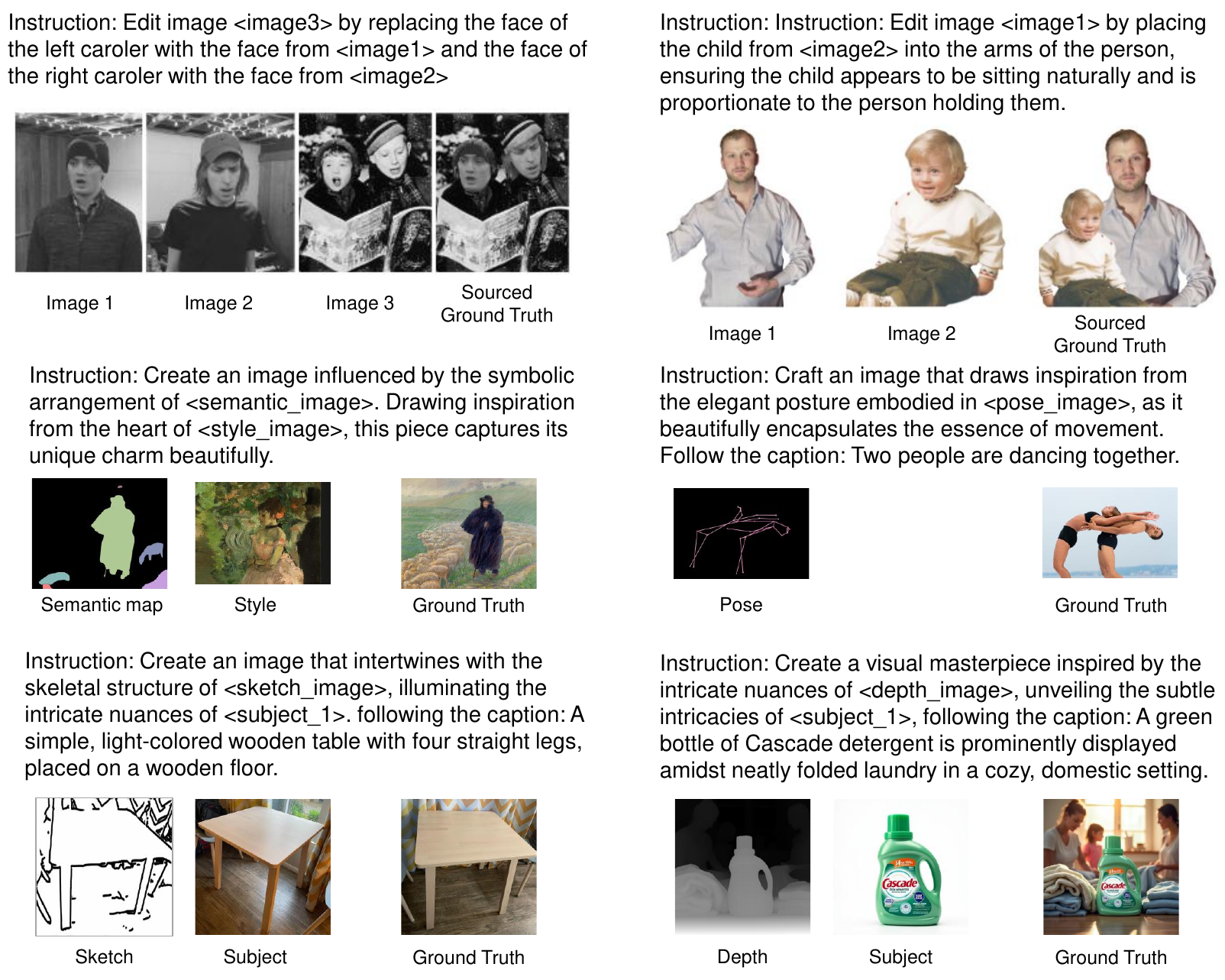}
    \vspace{-1em}
    \caption{Samples in \benchmark. The first row present real-world examples while others are from the synthetic part.}
    \vspace{-1em}
    \label{fig:benchmark-example}
\end{figure*}

\benchmark consists of 1,990 examples. The 1,000 examples from the real-world part represent real-world tasks sampled from the Reddit community \texttt{r/PhotoshopRequest}. 
The synthetic 990 examples are a test set split from \dataset generated using \engine. We show some examples in Figure \ref{fig:benchmark-example}.

\subsection{Real-world}
We collect 2,300 user queries from \texttt{r/PhotoshopRequest} community on Reddit, explicitly selecting tasks that require combining multiple input images to fulfill the requested edits. For each query, we gather all associated input images, the original text-based user instructions, and corresponding output images. To ensure data integrity and quality, each datapoint undergoes manual evaluation according to rigorous criteria. These criteria include verifying the necessity and appropriateness of each input image, assessing the logical coherence and relevance of instructions, and confirming accurate adherence to the instructions in the output image. In cases where multiple output images are provided for a single query, annotators select only one based on clarity, fidelity to the instruction, and overall quality.

% To handle noisy human instructions and clearly specify references to individual images, we employ GPT-4o to generate structured prompts and detailed editing instructions. The model is explicitly guided to closely adhere to the original user requests while systematically incorporating image reference tokens (e.g., \texttt{<image1>}) to indicate elements of the edit corresponding to specific input images. All VLM-generated instructions subsequently undergo manual review to ensure clarity, consistency, and conformity to a standardized meta-prompt format. In instances where GPT-4o omits references to one or more input images, annotators manually correct and add the appropriate image tokens.

% To provide insight into what edits are most commonly requested, we categorized each datapoint using the taxonomy structure proposed in OmniEdit~\cite{wei2024omnieditbuildingimageediting}. The taxonomy comprises five categories: Element Replacement, Element Addition, Style and Appearance Modifications, Spatial/Environment Modifications, and Attribute Transfer. Each datapoint was processed using GPT-4o~\cite{openai_gpt4o_2024}, following a standardized taxonomy prompt detailed in the Supplementary Material. 

% After applying these rigorous quality standards and review processes, 45\% of the collected data meet our criteria and are incorporated into the final benchmark dataset. This results in 1,000 examples, each comprising between two and six input images, a single structured instruction, and one output image as the golden answer.

\textbf{Taxonomy creation.}
We adopted the taxonomy structure introduced in OmniEdit \cite{wei2024omnieditbuildingimageediting} to categorize the types of edits represented in our benchmark. We utilized GPT-4o with the following prompt to generate the taxonomy for our dataset. The distribution of edit types in the real-world part is shown in Table~\ref{tab:example_distribution}.

To produce the meta-style prompts from noisy user instructions, we used the prompt with GPT-4o. We supplied all input images, the corresponding output image, as well as the original user instructions. 

\begin{table}[!t]
\centering
\caption{Distribution of examples across different categories in real-world samples.}
\label{tab:example_distribution}
\vspace{-1em}
\scalebox{0.88}{
\begin{tabular}{lr}
\toprule[1.5pt]
\textbf{Category} & \textbf{Num.} \\
\midrule
Element Replacement & 529 \\
Element Addition & 246 \\
Spatial/Environment Modifications & 111 \\
Attribute Transfer & 73 \\
Style and Appearance Modifications & 41 \\ \midrule
Total & 1,000 \\
\bottomrule[1.5pt]
\end{tabular}}
\vspace{-1em}
\end{table}

\begin{table*}[!t]
\centering
\caption{Evaluating MLLM-as-a-Judge in scoring with cross-validated human-annotated ground truth. GPT-4o and 4o-mini aligns closely with human scores in overall assessment. Human-Human shows the alignment between human annotators. }
\label{tab:eval_mllm_judge}
\vspace{-1em}
\resizebox{\textwidth}{!}{
\begin{tabular}{l|cccc|cccc|cccc}
\toprule[1.5pt]
\multirow{2}{*}{\textbf{Model}} & \multicolumn{4}{c|}{\textbf{Image Quality}} & \multicolumn{4}{c|}{\textbf{Instruction Following}} & \multicolumn{4}{c}{\textbf{Source Fidelity}} \\
% \cline{2-13}
 & Pearson & Spearman & MSE & MAE & Pearson & Spearman & MSE & MAE & Pearson & Spearman & MSE & MAE \\
\midrule
\multicolumn{13}{c}{\textcolor{violet!80!white}{\cellcolor{gray!10!white} \textbf{\textit{Realistic}}}} \\ \midrule 
Gemini-2.0-Flash & 0.385 & 0.403 & 2.220 & 1.118 & 0.422 & 0.447 & 2.750 & 1.216 & 0.354 & 0.356 & 3.747 & 1.409 \\
GPT-4o-mini & \textbf{0.466 }& \textbf{0.466} & \textbf{1.676} & \textbf{0.986} & 0.530 & 0.569 & 1.493 & 0.858 & 0.514 & \textbf{0.518} & \textbf{1.193} & \textbf{0.733} \\
GPT-4o & 0.432 & 0.420 & 2.486 & 1.223 & \textbf{0.624} & \textbf{0.616} & \textbf{1.405} & \textbf{0.764} & \textbf{0.613} & 0.513 & 1.216 & 0.736 \\
\rowcolor{gray!10!white} Human-Human & 0.589 & 0.573 & 1.611 & 0.936 & 0.665 & 0.590 & 1.152 & 0.720 & 0.571 & 0.441 & 1.473 & 0.824 \\ \midrule
\multicolumn{13}{c}{\textcolor{violet!80!white}{\cellcolor{gray!10!white} \textbf{\textit{Synthetic}}}} \\ \midrule
Gemini-2.0-Flash & 0.369 & 0.347 & 2.078 & 1.052 & 0.627 & 0.592 & 1.662 & 0.855 & 0.588 & 0.574 & 2.057 & 0.960 \\
GPT-4o-mini & \textbf{0.438} & \textbf{0.410} & \textbf{1.680} & \textbf{1.013} & 0.632 & 0.552 & \textbf{1.503} & 0.870 & 0.616 & 0.615 & 2.173 & 1.140 \\
GPT-4o & 0.406 & 0.374 & 2.350 & 1.083 & \textbf{0.668} & \textbf{0.608} & 1.537 & \textbf{0.843} & \textbf{0.659} & \textbf{0.626} & \textbf{1.573} & \textbf{0.860} \\
\rowcolor{gray!10!white} Human-Human & 0.629 & 0.648 & 1.823 & 0.930 & 0.721 & 0.735 & 1.820 & 0.867 & 0.694 & 0.708 & 1.840 & 0.840 \\
\bottomrule[1.5pt]
\end{tabular}
}
\vspace{-1em}
\end{table*}

\begin{tcolorbox}[breakable, prompt, title={Prompt of generating taxonomy for real-world queries}, segmentation style={draw=gray, line width=0.5mm}]
You are tasked with classifying image editing instructions into one of the following 5 categories:

1. Element Replacement
   - Face swaps
   - Object substitutions
   - Background replacements
   - Text replacements
   - Component swaps (wheels, screens, etc.)

2. Element Addition
   - Adding people to scenes
   - Adding objects to environments
   - Adding details or elements to objects
   - Adding text or graphics
   - Adding visual effects

3. Style and Appearance Modifications
   - Color adjustments
   - Lighting modifications
   - Artistic style transfers
   - Texture changes
   - Visual quality enhancements

4. Spatial/Environment Manipulations
   - Repositioning elements
   - Combining multiple images into layouts
   - Changing scale or proportion
   - Adjusting orientation or alignment
   - Creating composite images

5. Attribute Transfers
   - Transferring expressions between faces
   - Applying visual characteristics across images
   - Maintaining specific features while changing others
   - Matching visual properties (lighting, color)
   - Transferring specific details while preserving context

\vspace{-7pt}
\tcblower
\vspace{-7pt}

Given the following image editing instruction, classify it into exactly one of these 5 categories.
Respond with a JSON object with a single key "category" and the value being the category number (1-5).
\end{tcolorbox}

\begin{tcolorbox}[breakable, prompt, title={GPT-4o prompt for rewriting instructions}, segmentation style={solid}]

You are an expert at image editing. Your job is to write a prompt that would help machine learning models to edit images.
\newline

I'm showing you:
\newline
1. First, the INPUT IMAGE(S) that the user wants to edit.\newline
2. Then, the user's ORIGINAL INSTRUCTION (which might be noisy or unclear).\newline
3. Finally, the OUTPUT IMAGE after editing.\newline

Based on comparing these, please:\newline
1. Infer what specific edit was performed\newline
2. Write a clear, precise prompt that would help an AI model achieve this exact edit\newline

Your prompt should follow this format:

"Edit image \texttt{<image1>} by [specific editing instruction using clear terminology]"\newline

Here are some examples of good output prompts: \newline
- "Edit image \texttt{<image1>} by taking the person from \texttt{<image2>}, person from \texttt{<image3>} and adding them to \texttt{<image1>}."\newline
- "Edit image \texttt{<image2>} by transferring the background from \texttt{<image1>} and replacing the person with the person from \texttt{<image3>}"\newline
- "Edit image \texttt{<image1>} by faceswapping the person from \texttt{<image2>} into \texttt{<image1>}"\newline
\newline
Now, analyze the following:
\newline
ORIGINAL INSTRUCTION: \texttt{\{\{description\}\}}\newline

Please provide a well-structured, clear editing prompt that precisely describes the transformation shown in the images.

\end{tcolorbox}

\subsection{Synthetic}
Following established compatibility rules, we identified 33 different reference combinations. To ensure statistical robustness while maintaining a manageable evaluation scope, we selected 30 samples for each combination, resulting in a comprehensive set of 990 evaluation instances in the synthetic portion of \benchmark. Some examples are shown in row 2-3 in Figure \ref{fig:benchmark-example}.

% \twocolumn[{%
% \renewcommand\twocolumn[1][]{#1}%
% \centering
% \includegraphics[width=0.7\linewidth]{figure/real-world-benchmark-ex-small-cropped.pdf}
% \vspace{-1em}
% \captionof{figure}{We visualize example datapoints in the real-world half of \benchmark. These examples are sourced from Reddit's r/PhotoshopRequest community. }
% \label{fig:real-world-dataset-ex}
% }]

\subsection{Evaluation}
For overall assessment, we leverage MLLM-as-a-Judge using GPT-4o-mini. We validate the correlation of MLLM-as-a-Judge and human with a selected test set of 300 samples for either Realistic and Synthetic dataset. Our experiment in Table~\ref{tab:eval_mllm_judge} reveals that GPT-4o-mini surpasses other models in aligning with humans. Therefore, we use GPT-4o-mini for overall assessment.

\section{Details of Experiment}

\subsection{Model Settings}
In this section, we will introduce the hyper-parameters of image generative models to facilitate experiment reproducibility and transparency. All our experiments were conducted on a server equipped with two A800 and two 4090 GPUs.

\textbf{Open-source Unified Models.} We employed four open-source unified models. All hyper-parameters are detailed as follows:
\begin{itemize}[itemsep=0pt, leftmargin=*]
\item \textbf{OmniGen \cite{xiao2024omnigen}.} We set height=1024, width=1024, guidance\_scale=2.5, img\_guidance\_scale=1.6, seed=0 as default settings.
\item \textbf{ChatDit \cite{huang2024chatdit}.} We use the images-to-image API call provided in the GitHub.
\item  \textbf{ACE \cite{han2024ace}.} We use the ACE-0.6B-512px as ACE-chat model for multi-reference image generation in multi-turn. We set sampler='ddim', sample\_steps=20, guidance\_scale=4.5, guide\_rescale=0.5.
\item \textbf{Show-o \citep{xie2024show}.} We use multi-turn dialogue for multi-reference image generation. We set guidance\_scale=1.75, generation timesteps=18, temperature=0.7, resolution: $256\times256$.
\end{itemize} 
As reported in GitHub, Emu2-Gen \cite{sheynin2024emu} needs at least 75GB of memory. Due to the limitation of computation, it is not employed in our experiments.

\textbf{Other Models.} We utilize three proprietary models, GPT-4o, Claude-3.5-Sonnet, and Gemini-1.5-pro-latest as multimodal preceptors and Flux-dev, SD3, SD2.1 as image generators, with detailed settings as follows:
\begin{itemize}[itemsep=0pt, leftmargin=*]
    \item \textbf{Gemini-1.5-pro-latest \citep{geminiteam2023gemini}.} Temperature=1, top\_p= 0.95.
    \item \textbf{Claude-3.5-Sonnet \citep{anthropic2024claude35}.} Temperature=0.9.
    \item \textbf{GPT-4o \cite{openai_gpt4o_2024}.} Temperature=1, top\_p=1.
    \item \textbf{Flux1-dev \citep{Flux.1}.} guidance scale=3.5, num inference steps=50.
    \item \textbf{Stable Diffusion 3 \citep{esser2024scaling}.} guidance scale=7.0, num inference steps=28.
    \item \textbf{Stable Diffusion 2.1 \citep{rombach2022high}.} guidance scale=7.5, num inference steps=25.
\end{itemize}

% \subsection{Evaluation}
% For each reference, we compute quantitative metrics, as well as MLLM-as-a-judge for fair comparison.

% Given the complexity of free-form answers in image generation scenarios, the evaluation includes XXXX. 
\section{Additional Experiment Results}
\subsection{Full Results}
We present full results of model generation in Table \ref{tab: real+mllm-judge}. 

\begin{table*}[!ht]
  \centering
  % \ranjay{The metrics looks like the benchmark is already solved? Looks like there is hardly any room for improvement between what models can do and what ground truth is.}
  \setlength{\tabcolsep}{4pt} % Default value: 6pt
  \caption{Real-world image generation conditioned on multiple image references. Although today's image generative models produce high-quality outputs, most struggle with accurately following instructions and maintaining fidelity to source images. \textbf{IQ} - Image Quality, \textbf{IF} - Instruction Following, \textbf{SF} - Source Fidelity.}
  \label{tab: real+mllm-judge}
  \vspace{-1em}
  \resizebox{\textwidth}{!}{
  \begin{tabular}{l|ccc|ccc|ccc|ccc|ccc|ccc}
    \toprule[1.5pt]
    \multirow{2}{*}{\textbf{Model}} & \multicolumn{3}{c|}{\textbf{Element Add.}} & \multicolumn{3}{c|}{\textbf{Spatial Mani.}} & \multicolumn{3}{c|}{\textbf{Element Rep.}} & \multicolumn{3}{c|}{\textbf{Attribute Tran.}} & \multicolumn{3}{c|}{\textbf{Style Modi.}} & \multicolumn{3}{c}{\textbf{Overall}} \\
    & IQ & IF & SF & IQ & IF & SF & IQ & IF & SF & IQ & IF & SF & IQ & IF & SF & IQ & IF & SF \\
    \midrule
    \multicolumn{19}{c}{\textcolor{violet!80!white}{\cellcolor{gray!10!white} \textbf{\textit{Unified Model}}}} \\ \midrule 
    Show-o & 0.511 & 0.290 & 0.253 & 0.525 & 0.300 & 0.258 & \underline{0.508} & 0.268 & 0.240 & 0.548 & 0.301 & 0.260 & 0.473 & 0.307 & 0.259 & 0.513 & 0.293 & 0.254 \\
    OmniGen & \underline{0.553} & \underline{0.498} & \underline{0.429} & \underline{0.553} & \underline{0.461} & \underline{0.422} & 0.484 & \underline{0.450} & \underline{0.379} & \underline{0.567} & \underline{0.479} & \underline{0.408} & \underline{0.620} & \underline{0.590} & \underline{0.468} & \underline{0.555} & \underline{0.496} & \underline{0.421} \\
    ACE & 0.254 & 0.207 & 0.205 & 0.260 & 0.207 & 0.205 & 0.255 & 0.207 & 0.203 & 0.234 & 0.200 & 0.200 & 0.265 & 0.205 & 0.200 & 0.254 & 0.205 & 0.203 \\
    \midrule
    \multicolumn{19}{c}{\textcolor{violet!80!white}{\cellcolor{gray!10!white} \textbf{\textit{Compositional Framework}}}} \\ \midrule 
    ChatDiT & 0.629 & 0.390 & 0.345 & 0.643 & 0.411 & 0.352 & 0.643 & 0.434 & 0.360 & 0.682 & 0.466 & 0.395 & 0.688 & 0.522 & 0.424 & 0.657 & 0.445 & 0.375 \\
    Gemini+SD2.1 & 0.611 & 0.372 & 0.329 & 0.620 & 0.404 & 0.324 & 0.574 & 0.391 & 0.339 & 0.605 & 0.397 & 0.332 & 0.660 & 0.495 & 0.385 & 0.614 & 0.412 & 0.342 \\
    Claude+SD2.1 & 0.620 & 0.402 & 0.330 & 0.625 & 0.416 & 0.339 & 0.555 & 0.371 & 0.322 & 0.674 & 0.419 & 0.345 & 0.717 & 0.507 & 0.390 & 0.638 & 0.423 & 0.345 \\
    Gemini+SD3 & 0.764 & 0.590 & 0.478 & 0.729 & 0.589 & 0.453 & 0.725 & 0.540 & 0.452 & 0.715 & 0.556 & 0.452 & 0.785 & 0.640 & 0.485 & 0.744 & 0.583 & 0.464 \\
    Claude+SD3 & 0.744 & 0.578 & 0.454 & 0.751 & 0.586 & 0.456 & 0.675 & 0.497 & 0.408 & 0.745 & 0.556 & 0.441 & \textbf{0.795} & 0.629 & 0.478 & 0.742 & 0.569 & 0.447 \\
    Gemini+SD3.5 & \textbf{0.786} & \textbf{0.615} & \textbf{0.500} & 0.756 & 0.591 & \textbf{0.473} & \textbf{0.759 }& \textbf{0.558} & \textbf{0.459} & \textbf{0.789} & 0.564 & 0.441 & 0.780 & 0.610 & 0.460 & \textbf{0.774} & 0.588 & \textbf{0.467} \\
    Claude+SD3.5 & 0.767 & 0.563 & 0.469 & \textbf{0.777} & \textbf{0.598} & 0.472 & 0.700 & 0.506 & 0.406 & \textbf{0.789} & \textbf{0.625} & \textbf{0.466} & 0.790 & \textbf{0.654} & \textbf{0.498} & 0.765 & \textbf{0.589} & 0.462 \\ \midrule
    \rowcolor{gray!10!white} Ground Truth & 0.711 & 0.797 & 0.712 & 0.751 & 0.780 & 0.748 & 0.651 & 0.714 & 0.624 & 0.772 & 0.722 & 0.692 & 0.780 & 0.820 & 0.756 & 0.733 & 0.767 & 0.706 \\
    \bottomrule[1.5pt]
  \end{tabular}}
  \label{tab:omnigen_performance}
  \vspace{-1em}
\end{table*}

\subsection{More Visualizations}

\header{General Results.} 

\begin{figure*}
        \centering
    \includegraphics[width=0.85\linewidth]{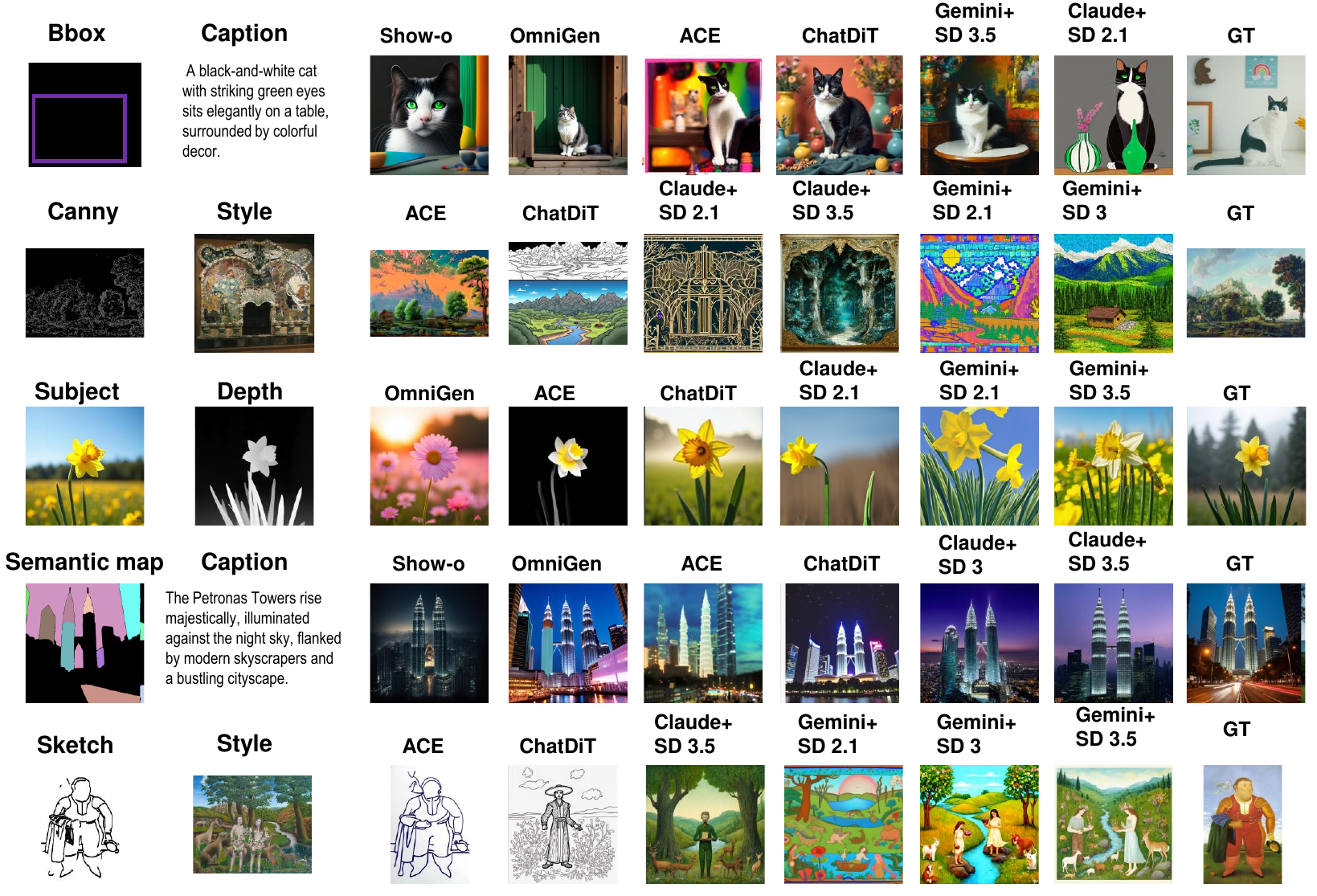}
    \caption{Image generation conditioned on a combination of two references.}
    \label{fig:vis2combo}
\end{figure*}

\begin{figure*}
        \centering
    \includegraphics[width=0.9\linewidth]{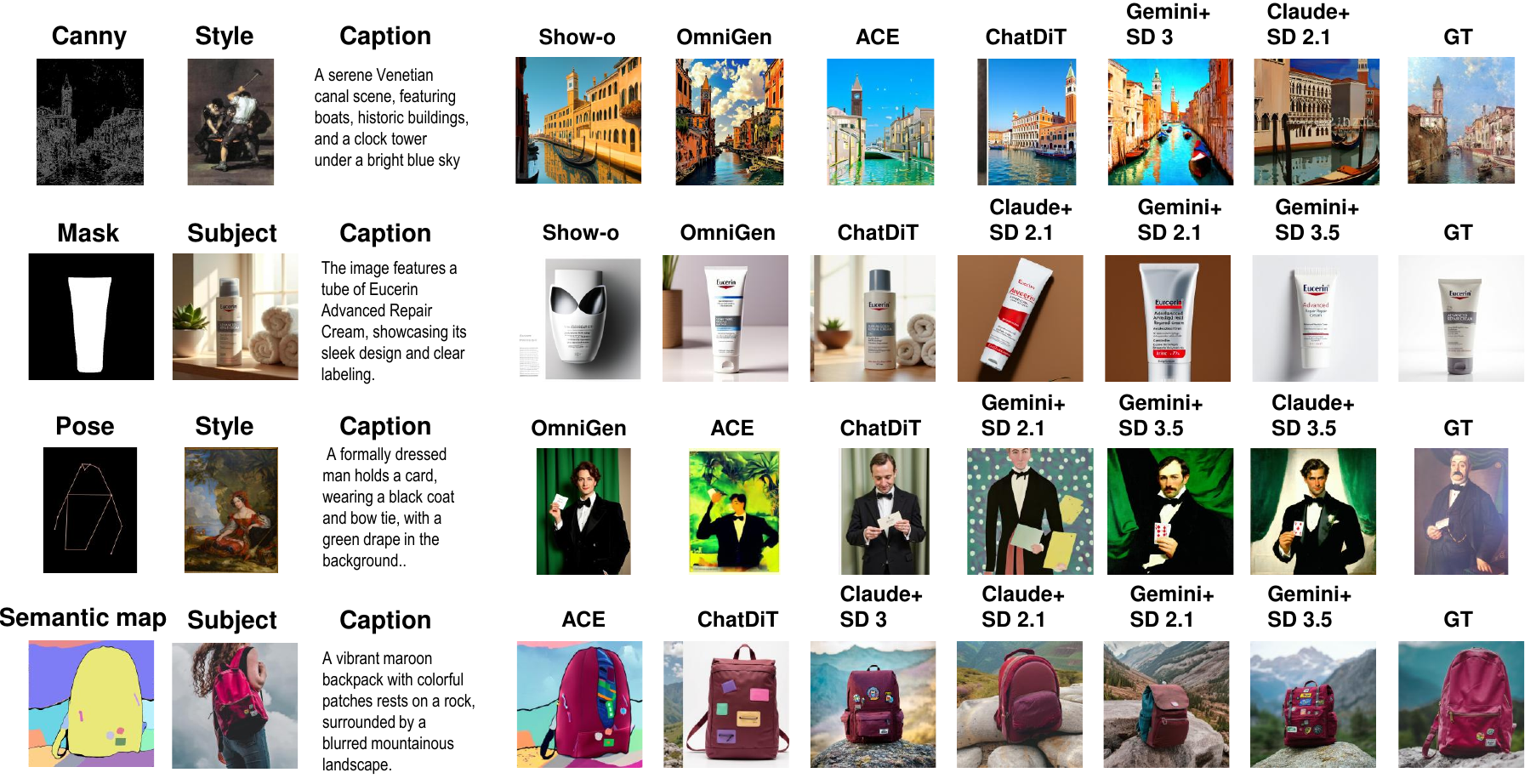}
    \caption{Image generation conditioned on a combination of three references.}
    \label{fig:vis3combo}
\end{figure*}

We show images generated by models under a combination of two references and three references in Figure~\ref{fig:vis2combo} and Figure~\ref{fig:vis3combo}, respectively. ``GT" means ground truth image.

Figure~\ref{fig:vis2combo} illustrates model performance when handling dual conditioning inputs. The results reveal significant variations in how different models interpret and integrate these multiple reference signals. Notably, most models perform poorly in maintaining the spatial accuracy of bounding boxes. For instance, ACE and ChatDiT inadequately follow sketch references, while compositional frameworks struggle with adherence to depth references. As an example, the Subject + Depth combination for flower generation, which tests the handling of both object appearance and spatial depth information, shows ACE maintaining depth relationships more effectively than other models. Encouragingly, all models demonstrate good performance in preserving object identity and art styles.

Figure~\ref{fig:vis3combo} extends this evaluation to three-reference scenarios, introducing more complex conditioning challenges. In the Canny + Style combination for Venetian canal scenes, for example, OmniGen and ACE exhibit better preservation of architectural details, whereas other models struggle with structural consistency. Furthermore, ACE demonstrates superior performance in spatial scene composition compared to other models, as evidenced by the Semantic map + Subject + Caption combination.


\begin{figure*}
\centering
\includegraphics[width=1\linewidth]{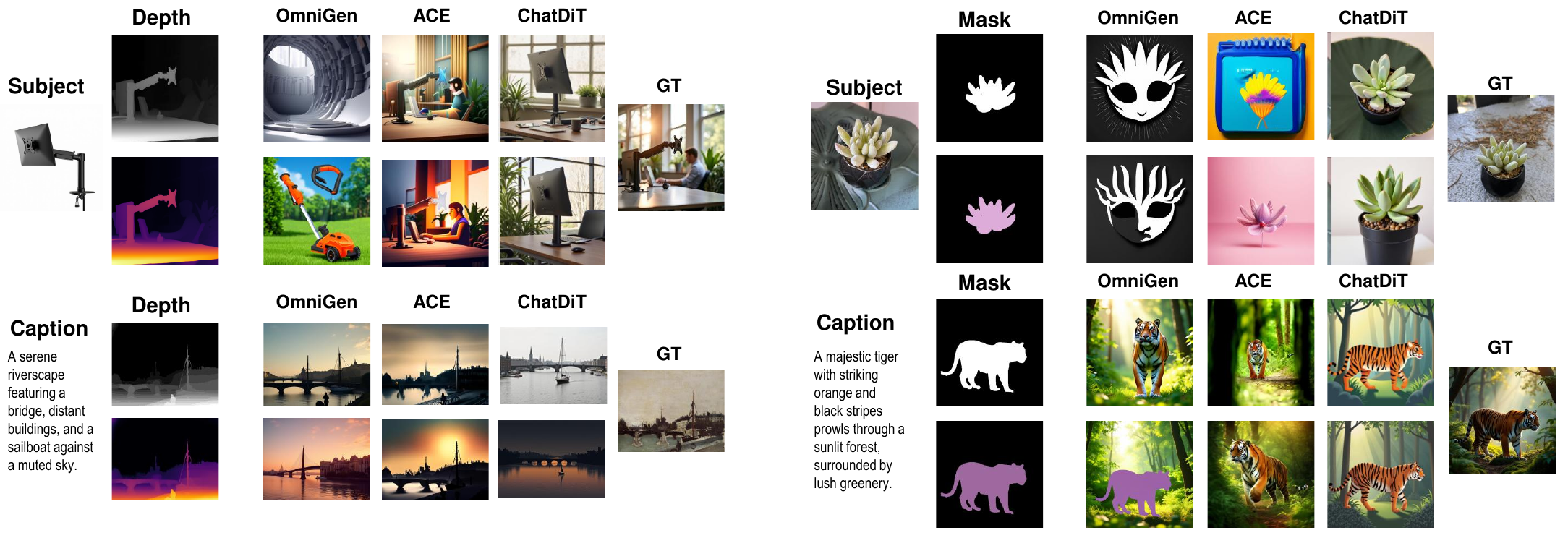}
\caption{Ablation study on depth map and mask conditioning formats. Left: Depth-conditioned generation results comparing model performance. Right: Mask-conditioned generation result. All models receive the same conditioning inputs but demonstrate varying adherence to spatial and semantic constraints.}
\label{fig:depth_mask_ablation}
\end{figure*}

% \begin{figure}
% \centering
% \includegraphics[width=1\linewidth]{figure/ablation_bbox.pdf}
% \caption{Ablation study on bounding box conditioning formats. We compare different models using bounding boxes in black and white backgrounds for conditional image generation. }
% \label{fig:bbox_ablation}
% \end{figure}

\header{Ablation study on reference formats.}

To evaluate the effectiveness of different conditioning formats, we conduct comprehensive ablation studies on depth map, mask, and bounding box inputs. Figure~\ref{fig:depth_mask_ablation} reveals distinct model behaviors under depth and mask conditioning. Depth-based conditioning (left panel) shows that models vary considerably in their ability to respect spatial depth relationships, with ACE can generate plausible scenes while OmniGen and ChatDiT struggle with geometric consistency. Mask-based conditioning (right panel) demonstrates each model's capacity for precise object placement and boundary adherence. ChatDiT shows more robust performance on different colors of mask references than others.

\begin{figure*}[t]
    \centering
    % 第一张子图
    \begin{subfigure}{0.49\linewidth}
        \centering
        \includegraphics[width=\linewidth]{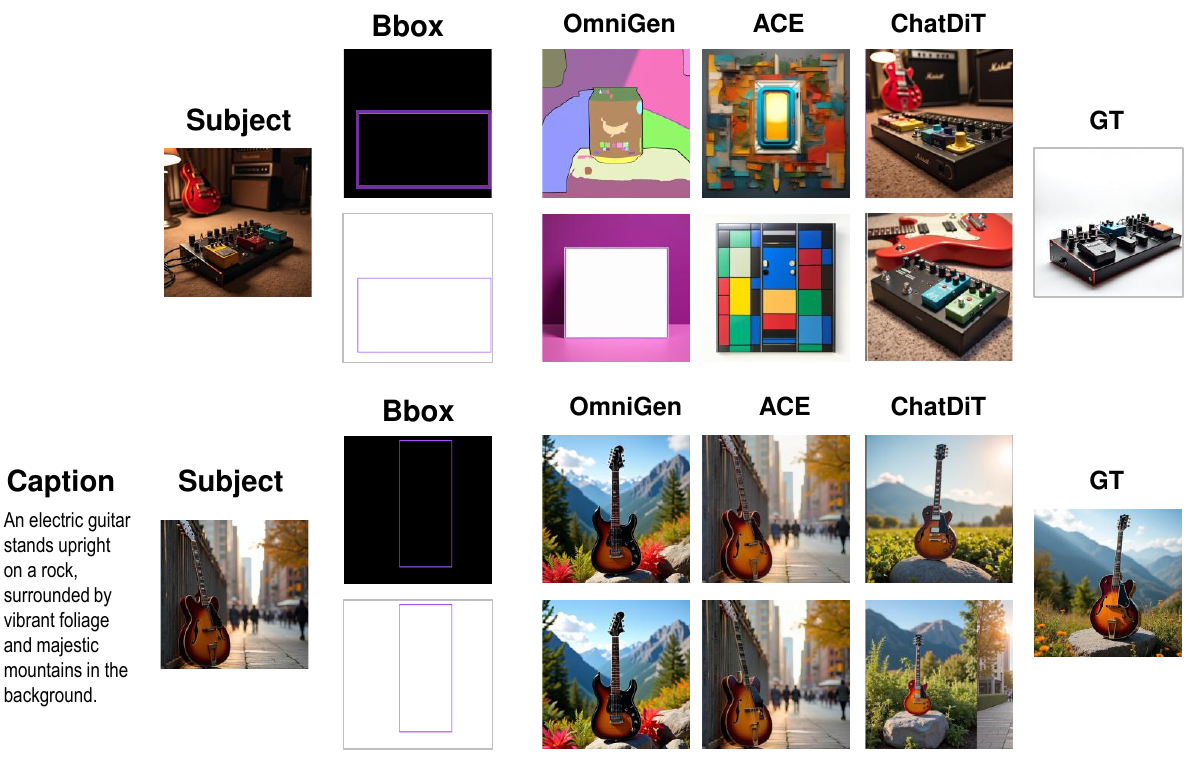}
        \caption{Ablation study on bounding box conditioning formats. We compare different models using bounding boxes in black and white backgrounds for conditional image generation.}
        \label{fig:bbox_ablation}
    \end{subfigure}
    \hfill
    % 第二张子图
    \begin{subfigure}{0.49\linewidth}
        \centering
        \includegraphics[width=\linewidth]{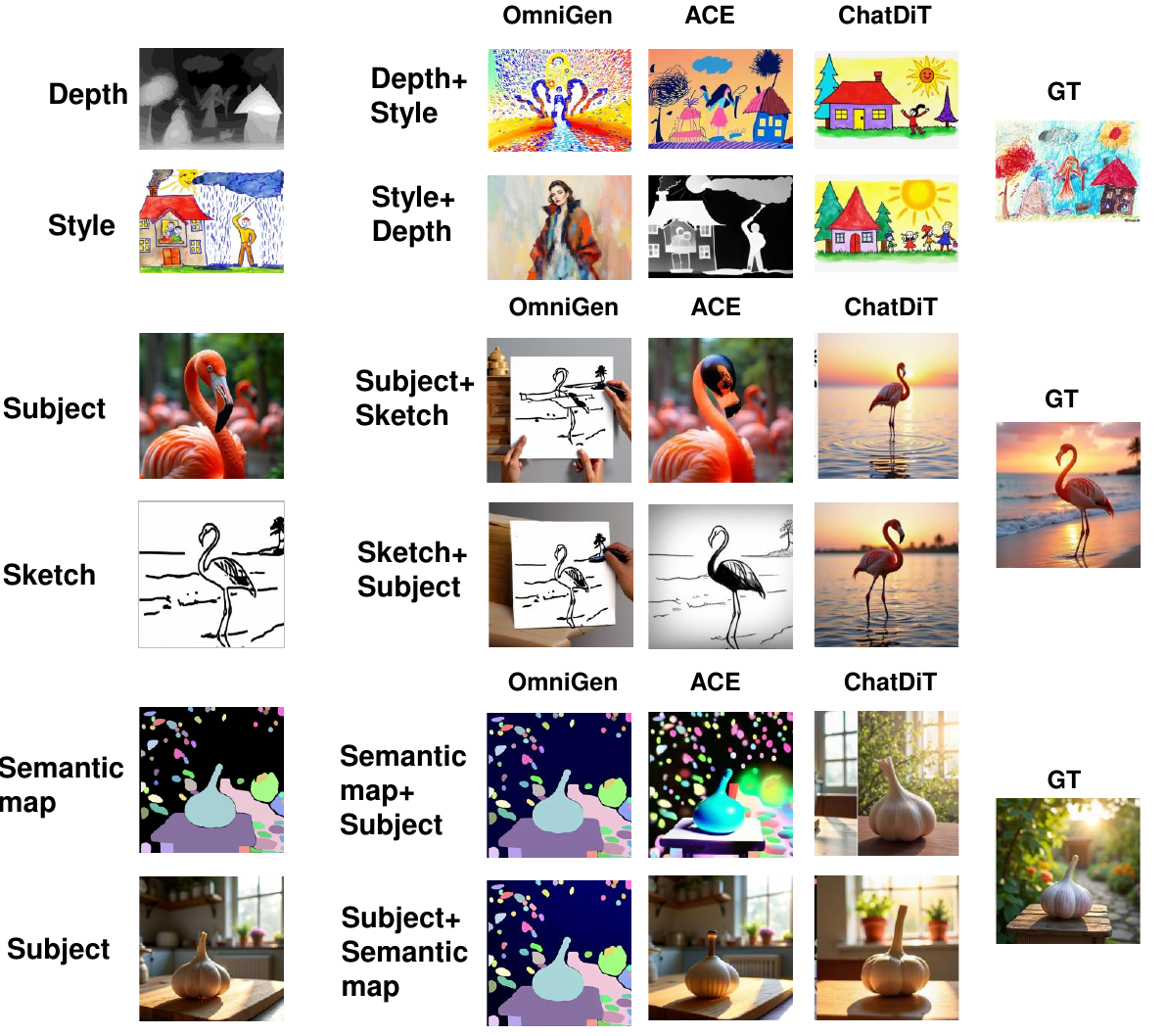}
        \caption{Ablation study on the impact of input reference order.}
        \label{fig:ablation_order}
    \end{subfigure}
    \caption{Ablation studies: (a) bounding box conditioning formats; (b) impact of input reference order.}
    \label{fig:ablation_combined}
\end{figure*}

Furthermore, Figure~\ref{fig:bbox_ablation} demonstrates that while all models can incorporate bounding box constraints, significant variations exist in spatial accuracy and semantic fidelity. ACE and ChatDiT exhibit different strengths in scene composition and detail preservation, but both perform poorly on spatial accuracy.

% \begin{figure}
%   \centering
%   \includegraphics[width=1\linewidth]{figure/ablation_order.pdf} % Adjust width as needed
%   \caption{Ablation study on the impact of input reference order.}
%   \label{fig:ablation_order}
% \end{figure}

\begin{figure*}
  \centering
  % Assuming your image file is named ablation_caption.pdf and is in the path
  \includegraphics[width=0.75\linewidth]{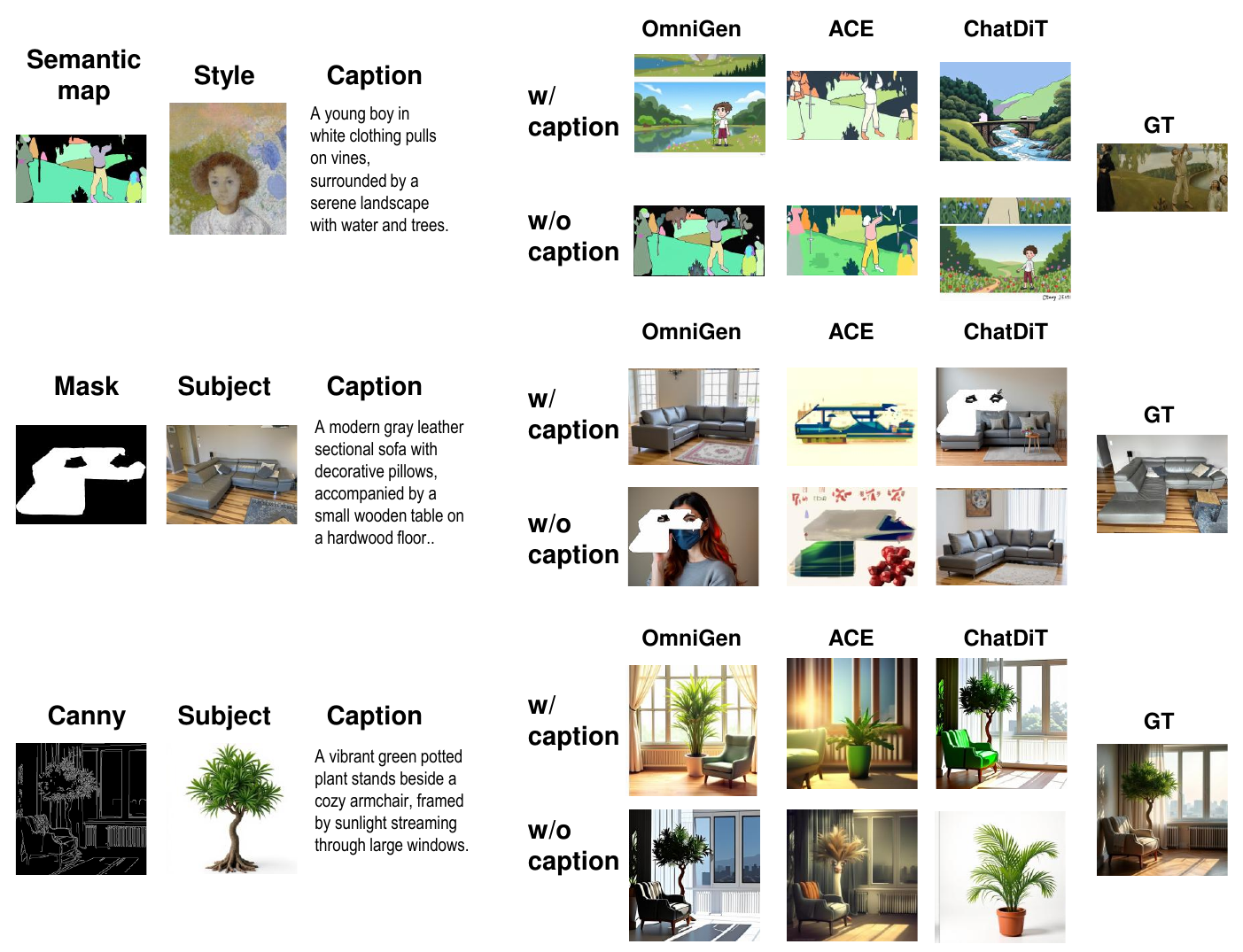} % Adjust width as needed
  \caption{Ablation study on the impact of captions. This figure illustrates the differences in image generation when a text caption is provided ("w/ caption") versus when it is omitted ("w/o caption").}
  \label{fig:ablation_caption}
\end{figure*}

\header{Ablation study on input image orders.}

Figure~\ref{fig:ablation_order} presents a grid of images demonstrating an ablation study on the effect of input order for different reference images. The results indicate that while all models can process combined visual inputs, the sequence in which these conditions are provided can noticeably influence the final output characteristics. For example, across OmniGen, ACE, and ChatDiT, variations in adherence to depth cues versus stylistic elements are observable when comparing ``Depth+Style" to ``Style+Depth" generations for the garlic scene. Similarly, the interplay between semantic maps and subject guidance, as shown in the bottom row of examples, yields distinct visual differences based on their ordering for all three models. This suggests that the sequential integration of features is not always commutative; the chosen order can be a significant factor in achieving the desired emphasis and balance between multiple visual constraints, affecting aspects like textural detail, spatial definition, and overall compositional fidelity.

\header{Ablation study on captions.}

Figure~\ref{fig:ablation_caption} highlights the significant impact of captions on the fidelity and semantic accuracy of generated images. When provided with a descriptive caption, models like OmniGen, ACE, and ChatDiT generally succeed in producing images that align well with the specified content and scene description, as demonstrated with the ``modern gray leather sectional sofa"  and the ``vibrant green potted plant stands beside a cozy armchair" examples. However, in the absence of captions (``w/o caption"), the generated outputs often exhibit a marked degradation in quality and relevance across all tested models. These uncaptioned results can range from abstract or distorted representations of the intended subject, as seen with OmniGen and ACE in the sofa examples, to the generation of entirely unrelated subject matter, notably where ChatDiT produced an image of a person wearing a mask instead of a sofa when no caption was provided for that scene. This underscores the crucial role of captions in guiding the models towards specific, coherent, and contextually appropriate image synthesis.

% \clearpage

% \bibliographystyle{ACM-Reference-Format}
% \bibliography{main}

% Identification of funding sources and other support, and thanks to
% individuals and groups that assisted in the research and the
% preparation of the work should be included in an acknowledgment
% section, which is placed just before the reference section in your
% document.

% This section has a special environment:
% \begin{verbatim}
%   \begin{acks}
%   ...
%   \end{acks}
% \end{verbatim}
% so that the information contained therein can be more easily collected
% during the article metadata extraction phase, and to ensure
% consistency in the spelling of the section heading.

% Authors should not prepare this section as a numbered or unnumbered {\verb|\section|}; please use the ``{\verb|acks|}'' environment.

% \section{Appendices}

% If your work needs an appendix, add it before the
% ``\verb|\end{document}|'' command at the conclusion of your source
% document.

% Start the appendix with the ``\verb|appendix|'' command:
% \begin{verbatim}
%   \appendix
% \end{verbatim}
% and note that in the appendix, sections are lettered, not
% numbered. This document has two appendices, demonstrating the section
% and subsection identification method.

\end{document}